\documentclass[letterpaper, 10 pt, conference]{ieeeconf}  

\IEEEoverridecommandlockouts                              
\usepackage{cite}
\usepackage{amsmath,amssymb,amsfonts}
\usepackage{mathabx}
\usepackage{algorithmic}
\usepackage{graphicx}
\graphicspath{ {./figures/} }
\usepackage{textcomp}
\usepackage{multirow}
\usepackage{hyperref}
\usepackage{comment}
\usepackage{caption}
\usepackage{subcaption}
\usepackage{array,multirow}
\usepackage{float}
\usepackage{xcite}
\usepackage[flushleft]{threeparttable}
\usepackage{xr}
\usepackage{wrapfig}
\makeatletter

\begin{document}
\bstctlcite{IEEEexample:BSTcontrol}
\title{Label Cleaning Multiple Instance Learning: Refining Coarse Annotations on Single Whole-Slide Image}
\author{Zhenzhen Wang,~Carla Saoud,~Sintawat Wangsiricharoen,~Aaron W. James,~Aleksander S. Popel,\\~Jeremias Sulam~\IEEEmembership{Member,~IEEE} 
\thanks{This work was supported in part by CISCO research grant CG\# 2686384, and NIH grants R01CA138264 and U01CA212007.}
\thanks{Z. Wang and J. Sulam (corresponding author) are with the Department of Biomedical Engineering, and the Mathematical Institute of Data Science, at Johns Hopkins University, MD 21218 USA (e-mail: \{zwang218,jsulam1\}@jhu.edu.)}
\thanks{A. S. Popel is with the Department of Biomedical Engineering at Johns Hopkins University, MD 21218 USA (e-mail: apopel@jhu.edu).}
\thanks{C. Saoud, S, Wangsiricharoen, and A. W. James are with the Department of Pathology at Johns Hopkins Medicine, MD 21218 USA (e-mail: \{csaoud1,swangsi1,awjames\}@jhmi.edu).}
}

\maketitle

\begin{abstract}
Annotating cancerous regions in whole-slide images (WSIs) of pathology samples plays a critical role in clinical diagnosis, biomedical research, and machine learning algorithms development. However, generating exhaustive and accurate annotations is labor-intensive, challenging, and costly. Drawing only coarse and approximate annotations is a much easier task, less costly, and it alleviates pathologists' workload. In this paper, we study the problem of refining these approximate annotations in digital pathology to obtain more accurate ones. Some previous works have explored obtaining machine learning models from these inaccurate annotations, but few of them tackle the refinement problem where the mislabeled regions should be explicitly identified and corrected, and all of them require a -- often very large -- number of training samples. We present a method, named Label Cleaning Multiple Instance Learning (LC-MIL), to refine coarse annotations on a \emph{single} WSI without the need of external training data. Patches cropped from a WSI with inaccurate labels are processed jointly within a multiple instance learning framework, 
mitigating their impact on the predictive model and refining the segmentation. Our experiments on a heterogeneous WSI set with breast cancer lymph node metastasis, liver cancer, and colorectal cancer samples show that LC-MIL significantly refines the coarse annotations, outperforming state-of-the-art alternatives, even while learning from a single slide. Moreover, we demonstrate how real annotations drawn by pathologists can be efficiently refined and improved by the proposed approach. All these results demonstrate that LC-MIL is a promising, light-weight tool to provide fine-grained annotations from coarsely annotated pathology sets. 

\end{abstract}

\section{Introduction}
\label{sec:introduction}
Pathology plays a critical role in modern medicine, and particularly in cancer care. Pathology examination and diagnosis on glass slides are the gold standard for cancer diagnosis and staging. In recent years, with advances in digital scanning technology, glass slides can be digitized and stored in digital form into whole-slide images (WSIs). These WSIs contain complete tissue sections and high-level morphological details, and are changing the workflow for pathologists \cite{Hanna2020,Melo2020}.

\begin{figure}[t]
\centerline{\includegraphics[width=\columnwidth]{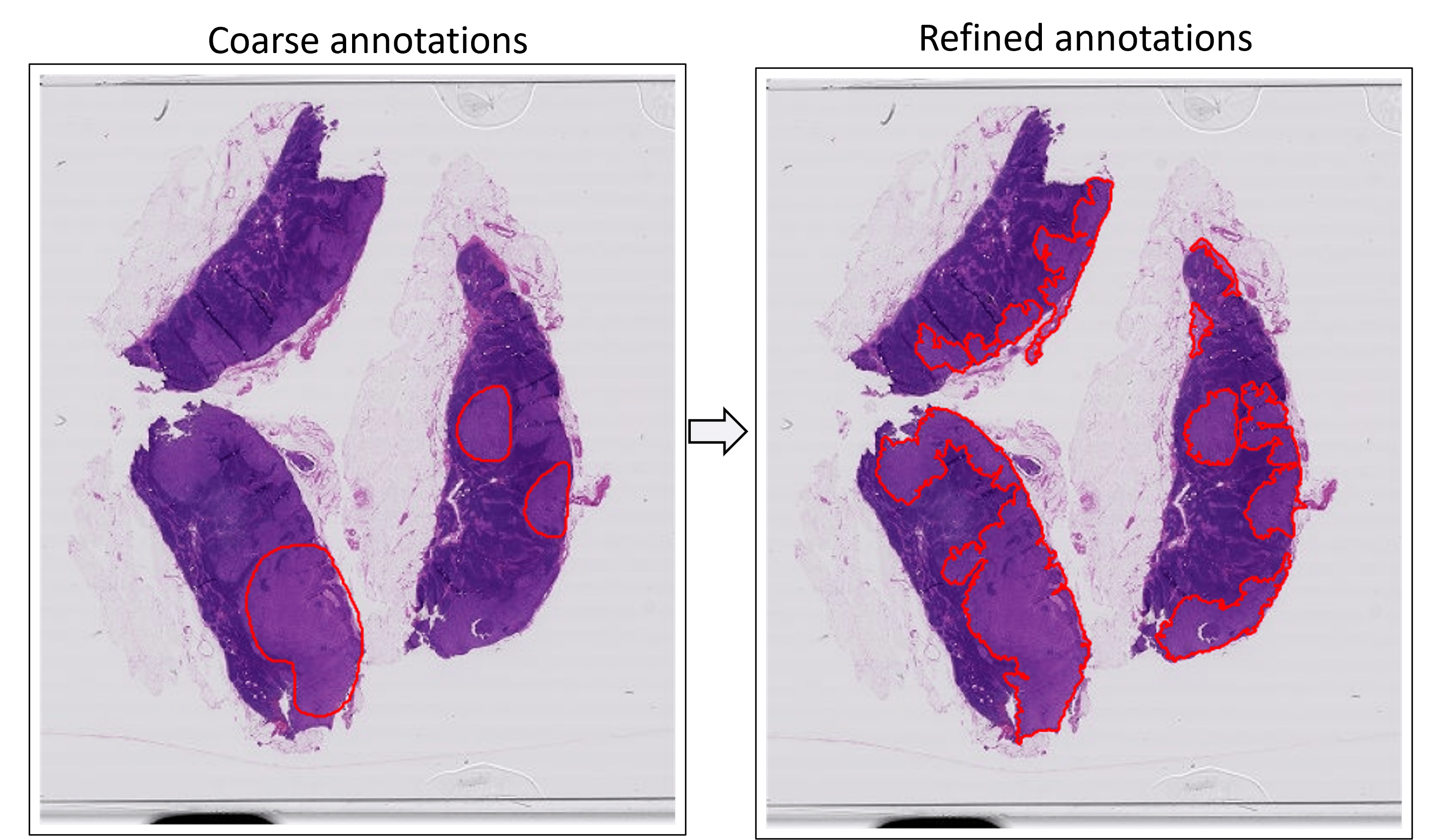}}
\caption{Example of the refining coarse annotation of breast cancer metastasis made by pathologists on a WSI of lymph node section, without requiring any external training data.}
\label{fig:eye-catching}
\end{figure}

Diagnosis by pathologists on a WSI typically include the description of cancer (\textit{i.e.}, presence, type, and grade of cancer), the estimation of tumor size, and observation of tumor margin (whether tumor cells appear at the edge of the tissue), which is important for planning therapy and estimating prognosis. Further information, such as the detailed localization of cancer, is usually not included in the routine pathology report. However, this local information on cancer is of great interest in biomedical and pharmaceutical research. The tumor microenvironment -- the stromal tissue and blood vessels surrounding the tumor cell clusters -- governs the tumor growth, response to treatment, and patient prognosis \cite{Song2015,Joyce2015}. The quantitative analysis of the tissue microenvironment, via downstream analysis of spatial statistics and other metrics, require the clear and accurate definition of the boundaries of the tumor, as in recent works  \cite{mi2020digital,Gong2019,Schwen2018}.

On the other hand, the local detection and segmentation of histopathology images is a significant and rapidly growing field in computational pathology. Early approaches mainly focused on extracting the morphological and texture features using image processing algorithms \cite{Gurcan2009}, including thresholding \cite{Petushi2006}, fuzzy c-means clustering \cite{Kande2010}, watershed algorithm \cite{Veta2013}, active contours \cite{Song2017}, among others. With the advent of artificial intelligence and deep learning, a number of deep neural network models have achieved encouraging results in biomedical image segmentation \cite{Ronneberger2015,Siddique2021}.

Whole-slide image segmentation, on the other hand, faces a unique challenge, since neural network models cannot be directly applied to the whole Gigapixel resolution WSI. A patch-based analysis (\textit{i.e.}, training and deploying a model on numerous small patches that are cropped from the WSI) is commonly used as an alternative, and supervised deep learning has been remarkably successful when deployed this way \cite{SWang2019}. However, high quantity (and quality) of fine-grained annotations are needed, including patch-level or pixel-level information. The latter is very costly to obtain, since detailed manual annotation on Gigapixel WSIs is extremely labor-intensive and time-consuming, and suffers from inter- and intra-observer variability \cite{Longacre2006, Foss2012, Loes2013}. For these reasons, state-of-art WSI sets with detailed annotations provided by expert pathologists are very limited. The lack of large datasets with detailed and trustworthy labels is one of the biggest challenges in the development and deployment of classical supervised deep learning models in digital pathology applications. 

Given the cost and difficulty of obtaining exhaustive and accurate annotations, a number of approaches attempt to address the segmentation problem in imperfect label settings \cite{Cheplygina2019}. Weakly-supervised learning aims to automatically infer patch-level (local) information using only slide-level (global) labels, but it typically requires thousands of WSIs as training samples \cite{Campanella2019,Lu2021,Srinidhi2021}. Semi-supervised learning, on the other hand, trains models on partially annotated WSI and makes predictions for the remaining unlabeled regions, yet the partial annotations must also be conducted by experts \cite{Cheng2020}. In this paper, we re-think the WSI annotation process from a more clinically applicable scenario: Drawing coarse annotations on WSIs (e.g., rough boundaries for the cancerous regions), is much easier than detailed annotations. Such coarse annotations need only similar effort and time as the slide-level labeling, and can even  be conducted by non-experts. Learning from those coarse annotations, and then refining them with computational methods, might be an efficient way to enrich the labeled pathology data with minimal effort. The resulting refined annotation can provide a more accurate ``draft'' for further detailed annotations by pathologists, thus alleviating their workload.
\begin{figure}[t]
\centerline{\includegraphics[width=\columnwidth]{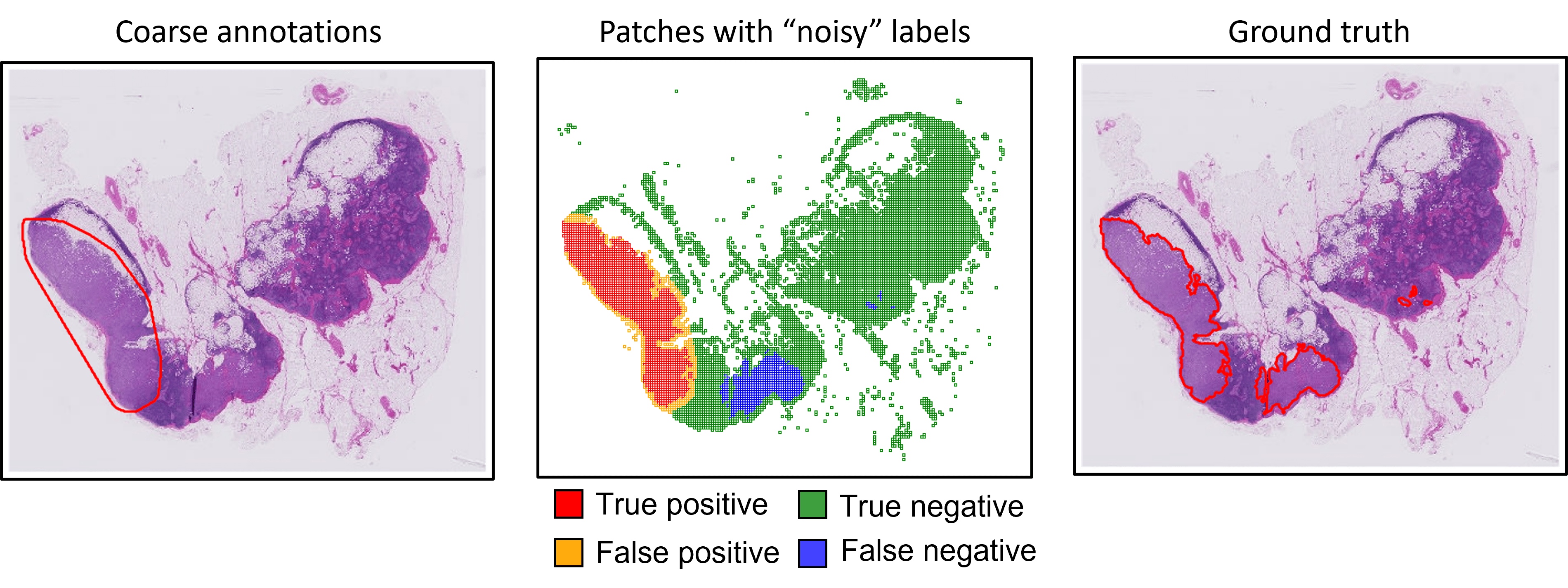}}
\caption{Illustration of the ``noisy'' label problem in coarse annotations of WSI.}
\label{fig:illustration_noisy_label}
\end{figure}

Our solution to the coarse annotation refinement problem is based on a multiple instance learning (MIL) framework that intrinsically incorporates the fact that the input annotations are imperfect, models this imperfection as patch-level label noise, and finally outputs a refined version of annotations by identifying the mislabeled patches and correcting them, as Fig. \ref{fig:eye-catching} illustrates. 
Our methodology, named Label Cleaning Multiple Instance Learning (LC-MIL), is validated in a heterogeneous set of 120 WSIs from three cancer types, including breast cancer metastasis in lymph nodes, liver cancer, and colorectal cancer, in both simulated and real experiments that demonstrate how the workload of pathologists can be alleviated. Importantly, and in order to make our method applicable to scenarios where just a few cases (slides) are available, our method can be trained and deployed on a single WSI, without the need for large training sets. Our approach significantly refines the original coarse annotations even while learning from a single WSI, and substantially outperforms state-of-the-art alternatives. To the best of our knowledge, this is the first time that multiple instance learning is employed to clean label noise, and ours is the first approach that allows for refining coarse annotations on single WSI.

The rest of the paper is organized as follows. In Section \ref{sec:RelatedWork} we provide an overview of related work. In Section \ref{sec:Methods}, we present the proposed methodology, LC-MIL, as well as other state-of-art baseline methods, in detail. We then proceed to our single-slide annotation refinement experiments in Section \ref{sec:experiments}. After that, we provide the implementation details in Section \ref{sec:implementation-details}, and discuss the sensitivity to parameters in Section \ref{sec:para-sens}. Lastly, we discuss and summarize both the implication and limitations of our approach in Section \ref{sec:Discussion} and Section \ref{sec:Conclusion}.

\section{Related Work}
\label{sec:RelatedWork}
In this section we discuss related work from the areas that are mostly related to our contribution: label noise handling and multiple instance learning in medical imaging.

Before we describe the specific techniques and studies, we first give a brief formulation of the problem to direct the reader to the appropriate context. The coarse annotations are considered as a  ``noisy'' label problem. More specifically, a WSI consists of a number of small patches, each of which can be assigned a label based on the annotations: patches within the annotated cancerous regions are assigned positive labels, and otherwise negative labels are assigned. However, not all of the labels are correct, given that the annotations are inaccurate. Comparing the coarse annotations with the ground truth, we will find some false positives (yellow) and false negatives (blue), as  Fig. \ref{fig:illustration_noisy_label} illustrates. Naturally, the model has no access to the ground truth, and aims at learning from, and modifying, the coarse annotations to retrieve the correct segmentation. This is the so-called ``annotation refinement'' problem, interpreted as identifying mislabeled samples and correcting their labels via machine learning approaches. This is a very challenging problem: the machine learning algorithm must be able to learn from inaccurate information and over considerable heterogeneity of tissue morphology and appearance, \emph{and} do so without an external collection of training data.

\subsection{Label noise handling}
\label{sec:label noise in machine learning}
We place ourselves in a machine learning scenario, where we are given a collection of training data -- pairs of samples and their corresponding labels, but part of training samples have their labels corrupted. The learning process, predicting a rule that assigns a label to a given sample, then become significantly more challenging. 
Label noise handling is an extensively researched problem in machine learning. A large family of approaches focus on enhancing the robustness of a machine system against label noise by designing sophisticated model architectures \cite{sukhbaatar2015training, Goldberger2017, Dgani2018}, choosing loss functions that are tolerant to label noise \cite{Ghosh2017, Matuszewski2018, rister2018ct}, and conducting ``label smoothing'' \cite{Gao2017, Pham2021, islam2021spatially}. This category of approaches typically does not evaluate the label accuracy or confidence of the training set, but tries to alleviate the impact of the corrupted labels in the performance of the model in a held-out test set.

Some other approaches attempt to evaluate the label accuracy based on predicted probabilities \cite{Northcutt2017,Ding2018,DBLP:journals/corr/abs-1906-11876} or loss values \cite{XWang2019, Shu2019}, under the intuition that samples with less confident predictions or unusually high loss values are more likely to be mislabeled. However, probability and loss cannot reveal the prediction confidence correctly in many poorly-calibrated models \cite{guo2017calibration}. An auxiliary set with clean labels --if available-- is popular in label noise detection, which is either used as a reference to identify potentially mislabeled samples \cite{Vo2015, Lee2018}, or re-weigh the training samples to mitigate the impact of mislabeled samples on the system \cite{Ren2018, Le2019, Mirikharaji2019}.

K-nearest neighbors (k-NN) based analysis \cite{Wilson1972}, on the other hand, can be used for ``editing'' corrupted labels without an auxiliary clean set. The basic idea is to discard samples that are not consistent with their $k$ nearest neighbors. In a recent work, k-NN has been deployed within deep learning frameworks, as the deep k-NN (DkNN) approach of \cite{Bahri2020}, which searches for neighbors in the feature space of a deep learning model. This method outperformed the state of the art for label noise correction, and we will revisit it in further detail as we describe our baseline in Section \ref{sec:baseline}.

\subsection{Multiple Instance Learning}\label{sec: MIL}
Unlike supervised learning settings, where each training sample comes with an associated label, in multiple instance learning (MIL), one only has labels associated to groups of samples, $X_i = \{x_k\}_{k=1}^{n_i}$, termed \emph{bags}, but not with the individual samples $x_i$, called \emph{instances}. Although individual labels $y_i$ exist for the instances $x_i$, they are unknown during training. However, the bag-level label $Y_i$ is a function of the instance-level labels $y_i$. This function was simply a \texttt{max} pooling operator when MIL was first proposed by Dietterich et al. \cite{DIETTERICH199731}, denoted as $Y = \underset{i}{\mathrm{max}}\{y_i\}$. 

Over the years, various alternative MIL formulations were developed. Ilse et al. \cite{Ilse2018} provide a generalization of MIL predictors as a composition of individual functions:
\begin{equation}\label{eq:MIL}
P_{Y=1}  = g(\sigma(f(X))).
\end{equation}
Here,  $f : \mathbb R^d \to \mathbb H$ is a transformation function  mapping individual instances to a feature space or label space; $\sigma : (\mathbb H)^k \to \mathbb H$ is a permutation invariant pooling function that aggregates the $k$ transformed instances within a bag; and $g : \mathbb H \to \mathcal Y$ finally maps  the aggregated instances to the corresponding bag label space.

When MIL is applied to medical imaging analysis, one typically considers an entire WSI as a bag, and regards patches cropped from the corresponding slide as instances. The bag label depends on the presence (positive) or absence (negative) of disease in the entire slide. While often successful in bag label prediction, these approaches require very large datasets with thousands of slides \cite{Campanella2019,Lu2021,Srinidhi2021}. Moreover, the disease localization, or instance-level predictions, usually suffers from the lack of supervision and underperforms the fully supervised counterparts \cite{Xu2014,Courtiol2018}. Moreover, local detection is usually considered as an additional -- sometimes optional -- task instead of the primary goal in MIL studies. Even if a heatmap or a saliency map is generated to highlight the diagnostically significant regions, the localization performance is not always quantitatively validated \cite{Campanella2019, Lu2021, Wang2018}.

Some works integrate other forms of weak annotations to the MIL framework to boost the local detection performance. CDWS-MIL, proposed by \cite{Jia2017}, introduced the percentage of the cancerous region within each image as an additional constraint to improve disease localization compared with using image-level labels only. CAMEL, proposed by \cite{Xu2019}, first splits the WSI into latticed patches, and considers each patch as a bag. A MIL model is then used to generate a pixel-wise heatmap for each patch. Those weak annotations, although easier to obtain than pixel-wise annotations, still need substantial effort and domain expertise. Our proposed method, on the other hand, is able to manage very coarse annotations, which are much easier to obtain, even without domain expertise.

\section{Methods}\label{sec:Methods}
In this section, we first formulate the problem at hand and then proceed to describe our proposed approach, label cleaning multiple instance learning (LC-MIL). Finally, we describe other state-of-art methods we used for comparison, including DkNN and Rank Pruning.
\subsection{Problem formulation}
\begin{figure}[h]
\centerline{\includegraphics[width=\columnwidth]{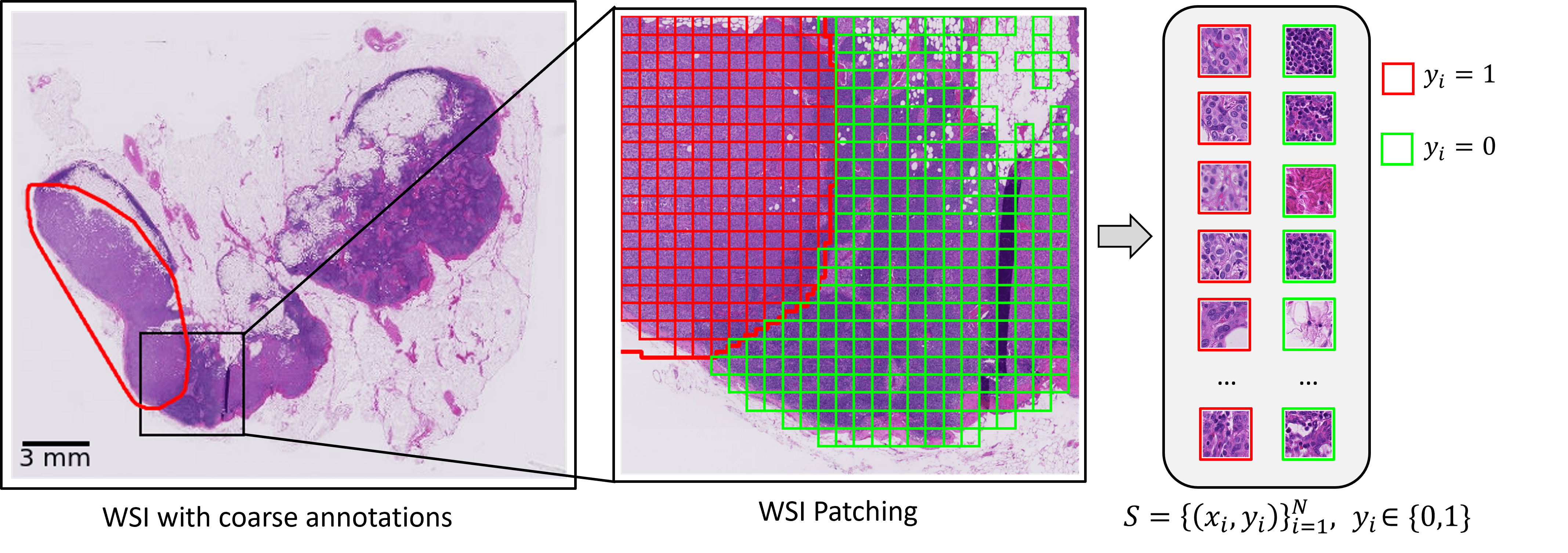}}
\caption{WSI patching and patch-level noisy labels assignment.}
\label{fig:patching}
\end{figure}

We consider a single WSI  with some coarse annotations for regions of interest (e.g., cancerous regions) as a dataset with noisy labels. The WSI is latticed to generate $N$ square patches, denoted as \(\mathbb{S} = \{(x_i,y_i)\}^{N}_{i=1}\). Each patch \(x_i\in\mathbb R^d\) is assigned a label \(y_i \in \{0,1\}\) based on the coarse annotations provided. To be more specific, a patch is assigned a positive label if it falls within the positively annotated area (\textit{e.g.}, cancerous region), and negative otherwise, as Fig. \ref{fig:patching} illustrates. In a practical setting, there may be some patches at the intersections of two classes of regions. For those patches, the assigned labels are decided by the location of patch centers. In other words, as soon as the center of a patch falls into  positively annotated regions, a positive label is assigned, and negative otherwise.

Importantly, we do not assume that all patches in the positively annotated area are actually positive: since the coarse annotations cannot delineate the disease regions precisely, there must be patches having an incorrect label assigned to them. Similarly, not all patches in the negative annotated area are true negatives, implying that there might be ``missed positive regions''. The problem of refining this coarse (and inaccurate) annotation can then be interpreted as detecting the mislabeled patches.

\subsection{Label Cleaning MIL (LC-MIL)}\label{sec:method-LC-MIL}
\begin{figure}[t]
\centerline{\includegraphics[width=\columnwidth]{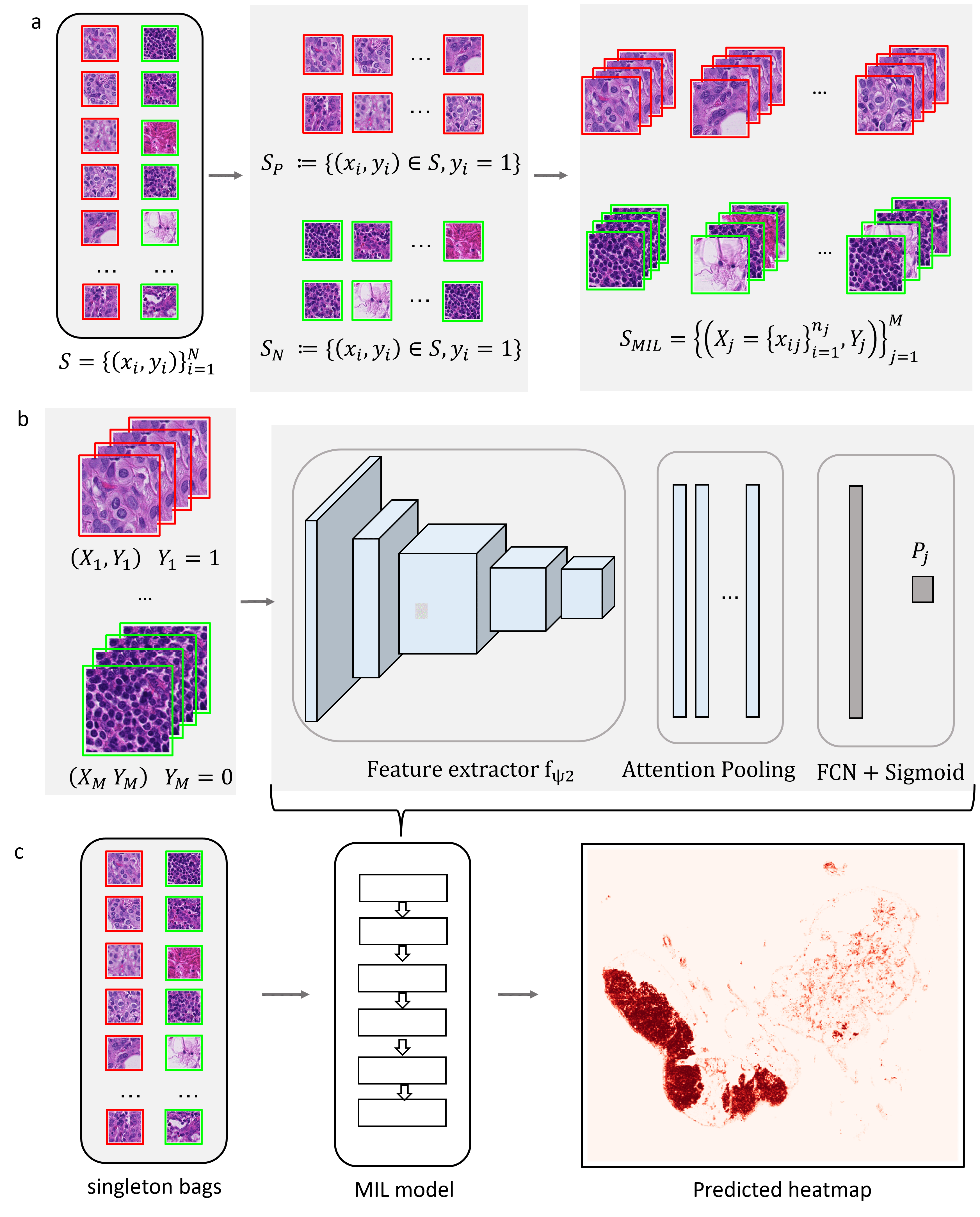}}
\caption{\textbf{Overview of the LC-MIL framework.} \textbf{(a)} MIL dataset construction: patches in WSI are split into two subsets \(\mathbb{S}_P\) and \(\mathbb{S}_N\) based on the original coarse annotations. \textbf{(b)} An attention based MIL model is trained on bags \(X_i = \{ x_{ij}\}_{j=1}^{n_i}\). \textbf{(c)} At inference, each patch is considered as a singleton bag and receives risk score to generate a predicted heatmap.}
\label{fig:methodology-LC-MIL}
\end{figure}

Our proposed methodology, Label Cleaning Multiple Instance Learning (LC-MIL), tackles the noisy label problem from a MIL framework. Broadly speaking, our approach consists on a multiple instance learning model that is trained to classify \emph{bags} of patches from a single WSI image with noisy labels. Once this model is trained, it is then employed to re-classify all patches in the image by constructing singleton bags. The corresponding predictions are used to correct the original noisy labels, and in doing so, refining the inaccurate segmentation. Our approach is general with respect to the specifics of the multiple instance model, and different specific methods are possible. In this work we focus on an implementation of based on a deep attention mechanism \cite{Ilse2018}, as well as an alternative based  neural network pooling \cite{wang2018revisiting}.  
Fig. \ref{fig:methodology-LC-MIL} depicts an overview of out LC-MIL algorithm (in its attention-based implementation), and we now expand on the details of each component. 

\textbf{MIL dataset construction:} As briefly mentioned above, in order to give our approach maximal flexibility in terms of the availability of training data, we situate ourselves in the case of having a single WSI, which represents all the data available to the algorithm -- we will comment on extensions to cases with access to larger datasets later. 

From a single WSI, we construct our noisy dataset of image patches \(\mathbb{S} = \{(x_i,y_i)\} ^{N}_{i=1}\), comprising positive and negative cases: (1) \(\mathbb{S}_P := \{(x_i, y_i) \in \mathbb{S} : y_i = 1\}\), which consists of all of the patches with a  positive label; (2) \(\mathbb{S}_N := \{(x_i, y_i) \in \mathbb{S} : y_i = 0\}\), which consists of all of the patches with a negative label. In our context, where labels are determined by only coarse and inaccurate annotations, there are likely false positives in \(\mathbb{S_P}\) and false negatives in \(\mathbb{S_N}\). However, we assume that the majority of the labels in both subsets are correct. With this assumption, we construct a positive bag $X_j$ by selecting (uniformly at random) $n_j$ instances from \(\mathbb{S}_P\), and construct a negative bag analogously with instances from \(\mathbb{S}_N\). We are able to create virtually as many bags as desired from one single WSI, since the sampling is conducted with replacement 
The constructed MIL dataset with \(M\) bags can be defined as:
\begin{equation} 
    S_\text{MIL} =  \{(X_j = \{ x_{ij}\}_{i=1}^{n_j}, Y_j)  \}_{j=1}^{M},
\label{eq:MIL set}
\end{equation} 
where \(x_{ij}\) is the \(i\)th instance in \(X_j\), and \(Y_j\) refers to the bag-level label, given by
\begin{equation}
    Y_j = \begin{cases}
                1, & \text{if } X_j  \subset \mathbb{S}_P \\
                0, & \text{if } X_j  \subset \mathbb{S}_N.
            \end{cases}
\end{equation}
The number of bags $M$, and the number of instances within a bag $n_j$ are adjustable parameters, which we gave the detailed setting and discussed the sensitivity of different choices in Section \ref{sec:para-sens}.

\textbf{MIL predictor:} We built a deep MIL model to predict the bag-level score \(P_j\), under the composite function framework as described in Section \ref{sec: MIL} and Eq. \eqref{eq:MIL}. Our approach is general in that it admits different forms for the functions $g, \sigma$ and $f$, leading to different MIL methods. We focus first on a predictor similar to that the one proposed in \cite{Ilse2018}, composed of the three functions below:
\begin{enumerate}
    \item A feature extractor, parameterized by a deep neural network \(f_{\psi}\), is used to map the instance \(x_{ij}\) to a low-dimensional embedding, denoted as:
        \begin{equation}
        h_{ij} = f_{\psi}(x_{ij}). 
        \end{equation}

    \item An attention-based MIL pooling operator $\sigma$, which gives each instance a weight \(w_{ij}\) and aggregates the low-dimensional embeddings of all instances using a weighted average to generate a representation for the whole bag  \cite{Ilse2018}. Importantly, these weights are learnable functions of the features, too. A softmax function is used to re-scale weights so that they all lie in the range [0,1], and sum up to 1. More specifically, we have that:
        \begin{equation}
        z_j = \sigma(\{h_{ij}\}_{i=1}^{n_j}) = \sum^{n_j}_{j=1}{w_{ij}h_{ij}},
        \end{equation}
    with weights given by
    \begin{equation}
        w_{ij} = \frac{\exp\{W \tanh{(Vh_{ij})}\}}{\sum^{n_j}_{k=1}\exp\{W \tanh{(Vh_{ik})}\}}.
    \end{equation}
    Here, both \(W\) and \(V\) are learnable parameters.

    \item A linear classifier $g$, which predicts the bag label based on the computed representation, $z_i$. The predicted score, $P_i$, is finally obtained by the appropriate logistic function, defined as:
    \begin{equation}
    	P_j = g(z_j) = \frac{1}{1+e^{-\langle g,z_j \rangle}} ,
    \end{equation}
    where $g$ is a learnable vector.
\end{enumerate}

Alternatively, and for the purpose of understanding the implications of the choice in the MIL predictor, we also consider 
a second implementation of LC-MIL by using another MIL formulation called mi-Net \cite{wang2018revisiting}, whose formulation can be formally expressed as the follows. Given a bag of instances, $X_j = \{x_{ij}\}_{i=1}^{n_j}$, the bag-level likelihood $P_j$ is given by:
\begin{equation}
    P_j = \frac{1}{n_j}\sum_{i=1}^{n_j} f(x_{ij}), 
\end{equation}
where $f(x_{ij})$ is a instance-level classifier parameterized by a deep neural network. To differentiate these two instantiations of our approach, we name them as \emph{LC-MIL-atten} and \emph{LC-MIL-miNet}, respectively.

Unlike \cite{Ilse2018}, who employ a cross-entropy loss, we use the focal loss \cite{mukhoti2020calibrating} to train our MIL predictors in order to promote better calibration of the predicted probabilities. This loss is defined as:
\begin{equation}
    L = -(Y_j (1-P_j)^{\gamma}\log{P_j} + (1-Y_j )P_j^{\gamma}\log{(1-P_j)} ).
\label{eq:focal-loss}
\end{equation}
 We set the parameter \(\gamma\) as suggested by \cite{mukhoti2020calibrating}. That is \(\gamma = 5\) for \(P(Y_j = 1) \in [0,0.2)\), and \(\gamma = 3\) for \(P(Y_j = 1) \in [0.2,1]\).

\textbf{Inference:} There is typically a ``gap'' between bag-level and instance-level prediction in MIL approaches, where the instance-level score is not directly predicted by the model. Although the attention pooling operator provides a way to locate key instances, those weights cannot be interpreted as instance-level scores directly. Here, we propose to use ``singleton bags'' as a simple solution, enabling us to exploit the (calibrated) scores provided by the  classification model: for bags consisting of only one instance, the bag-level prediction is equal to the instance-level prediction. In this way, during the inference phase, we revisit the noisy dataset  \(\mathbb{S} = \{(x_i,y_i)\} ^{N}_{i=1}\) using the trained MIL model. Instead of randomly choosing a subset of instances to pack a bag, as done during training,  every single instance \(x_i\) is now considered as a ``single-instance bag''. Since there is only one instance in a bag, the attention-based pooling in the MIL framework has no impact during inference -- in other words, the attention weights are always set to 1 for any instance -- and the predicted score for each instance can be denoted as:
\begin{equation}
    \hat{p_i} = \frac{1}{1+e^{-\langle g, f_{\psi_2}(x_{i})\rangle}}.
\end{equation}
A predicted heatmap is then generated based on these likelihoods, where the value of each pixel refers to the risk score \(p_i \in [0,1] \) of the corresponding patch in the WSI. 


In addition to the central machine learning component of our method, detailed above, our experiments involve other implementations details, which we detail in Section \ref{sec:implementation-details}. All software implementations of our methods are publicly available\footnote{Available at \url{https://github.com/Sulam-Group/MIL-pathology}}.

\subsection{Comparative methods}\label{sec:baseline}
It should be noted that, to the best of our knowledge, the problem of learning a segmentation algorithm for WSI data from a single slide with inaccurate annotations has never been studied before. As a result, there are no available methods that can be deployed in an off-the-shelf manner. Thus, to provide comparison methods for our problem, we adapt state-of-the-art algorithms to our setting, where label noise should be corrected without an auxiliary clean set.

\noindent\textbf{Deep k-nearest neighborhood (DkNN)} \cite{Bahri2020}: Deep k-nearest neighborhood detects label noise using the assumption that instances within the same class should cluster together in the feature space. A feature extractor parameterized by a preliminary deep neural network is used to map instances to the feature space. In the feature space, instances having the same labels with their neighbors' are trusted, while those having inconsistent labels with neighbors are more likely to be mislabeled. These suspicious instances are then relabeled via majority voting. This re-labeling procedure is conducted using a standard k-NN classifier. 

\noindent\textbf{Rank Pruning} \cite{northcutt2017learning}: Rank Pruning identifies label noise using the prediction confidence in a two-step process, where the instances with low prediction confidence are removed in the second-round training. To be more specific, a binary classifier is first fitted on the noisy-labeled instances in a cross-validation manner. Each instance is then given a preliminary probability score, which are considered to reveal the prediction confidence. The types of instances are considered as un-trusted and removed, including those getting abnormally high probability scores while the original noisy labels are negative, and the those getting abnormally low probability scores while the original noisy labels are positive. After pruning the un-trusted instances, the classifier is then re-fitted on the remaining, and a new probability score for each instance produced by this updated classifier trained on trusted samples.
\\

We also list the two implementations of LC-MIL that are used in the experiments for clarity. As we described in Section \ref{sec:method-LC-MIL}, these two implementations share all the components of the LC-MIL framework, except the MIL predictor.
\begin{enumerate}
    \item \textbf{LC-MIL-atten}: It uses the attention-based MIL \cite{Ilse2018} as the MIL predictor, which aggregate embedding of instances using an attention mechanism.
    \item \textbf{LC-MIL-miNet}: It uses an alternative MIL formulation called mi-Net \cite{wang2018revisiting}, which aggregates results of the instance-level classifiers using a mean-pooling operator.
\end{enumerate}

\section{Experiments} \label{sec:experiments}
In this section, we first describe our dataset, and then show two coarse annotation refinement scenarios in simulated and real-world settings, separately.
\subsection{Dataset}
We evaluated the coarse annotation refinement performance on three publicly available histopathology datasets. These datasets were chosen because they include different tissues and degrees of morphological heterogeneity, but also because of the availability of expert annotations that will be regarded as ground-truth for the quantitative evaluation of our method.

\begin{enumerate}
\item CAMELYON16\cite{Litjens2018}: contains a total of 399 hematoxylin and eosin (H\&E) stained WSIs of lymph node sections from breast cancer patients. Detailed hand-drawn contours for metastases are provided by expert pathologists.
\item PAIP2019\cite{Kim2021}: contains a total of 100 H\&E stained WSIs of liver cancer resection samples. The boundary of viable tumor nests was annotated precisely by expert pathologists. The viable tumor nests annotations are available for 60 WSIs. 
\item PAIP2020\footnote{De-identified pathology images and annotations used in this research were prepared and provided by the Seoul National University Hospital by a grant of the Korea Health Technology R\&D Project through the Korea Health Industry Development Institute (KHIDI), funded by the Ministry of Health \& Welfare, Republic of Korea (grant number: HI18C0316).}: contains a total of 118 H\&E stained WSIs of colorectal cancer resection samples. The contours of the whole tumor area, which is defined as boundary enclosing dispersed viable tumor cell nests, necrosis, and peri- and intratumoral stromal tissues, are provided by expert pathologists. The whole tumor annotations are available for 47 WSIs. 
\end{enumerate}

\begin{figure}[t]
\centerline{\includegraphics[width=\columnwidth]{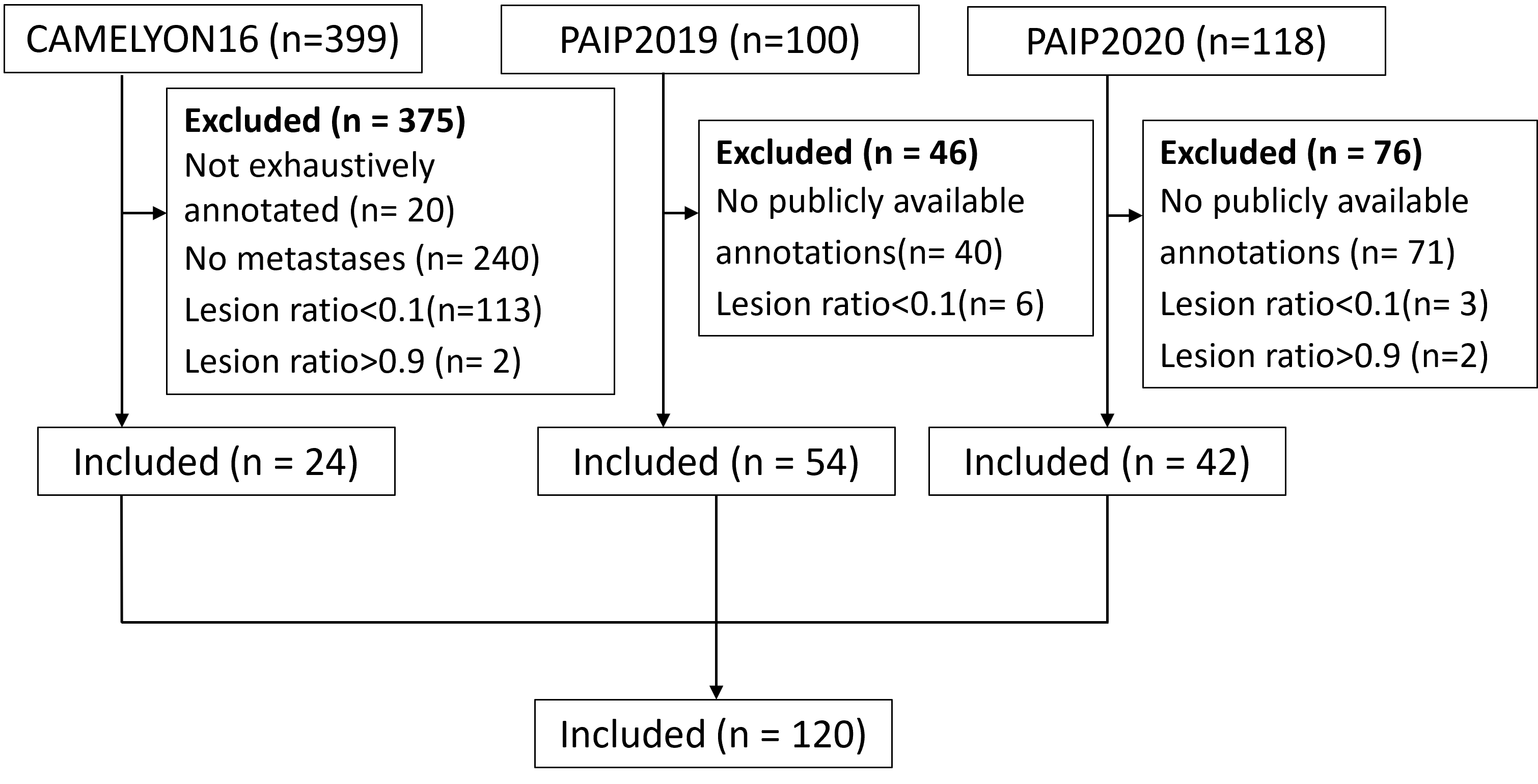}}
\caption{\textbf{Data inclusion diagram.} A total of 120 slides are included. 24 of them are lymph node sections from breast cancer patients (CAMELYON16), 54 of them are  liver cancer resection samples (PAIP2019), and the other 42 are colorectal cancer resection samples (PAIP2020).}
\label{fig:data_inclusion}
\end{figure}

The experimental design had the goal of validating the segmentation refinement, capability of our method on every single slide independently, and thus slides with almost all regions of a single class (cancer or normal) were excluded to ensure that our model had sufficient positive and negative samples. We set an upper bound of 90\% and a lower bound of 10\% for the ratio of lesion area as the data inclusion criteria. Slides marked with ``not exhaustively annotated'' were also excluded. A total of 120 slides (CAMELYON16: 24; PAIP2019: 54; PAIP2020: 42) were included, as shown in Fig. \ref{fig:data_inclusion}.

\begin{figure}[t]
\centerline{\includegraphics[width=\columnwidth]{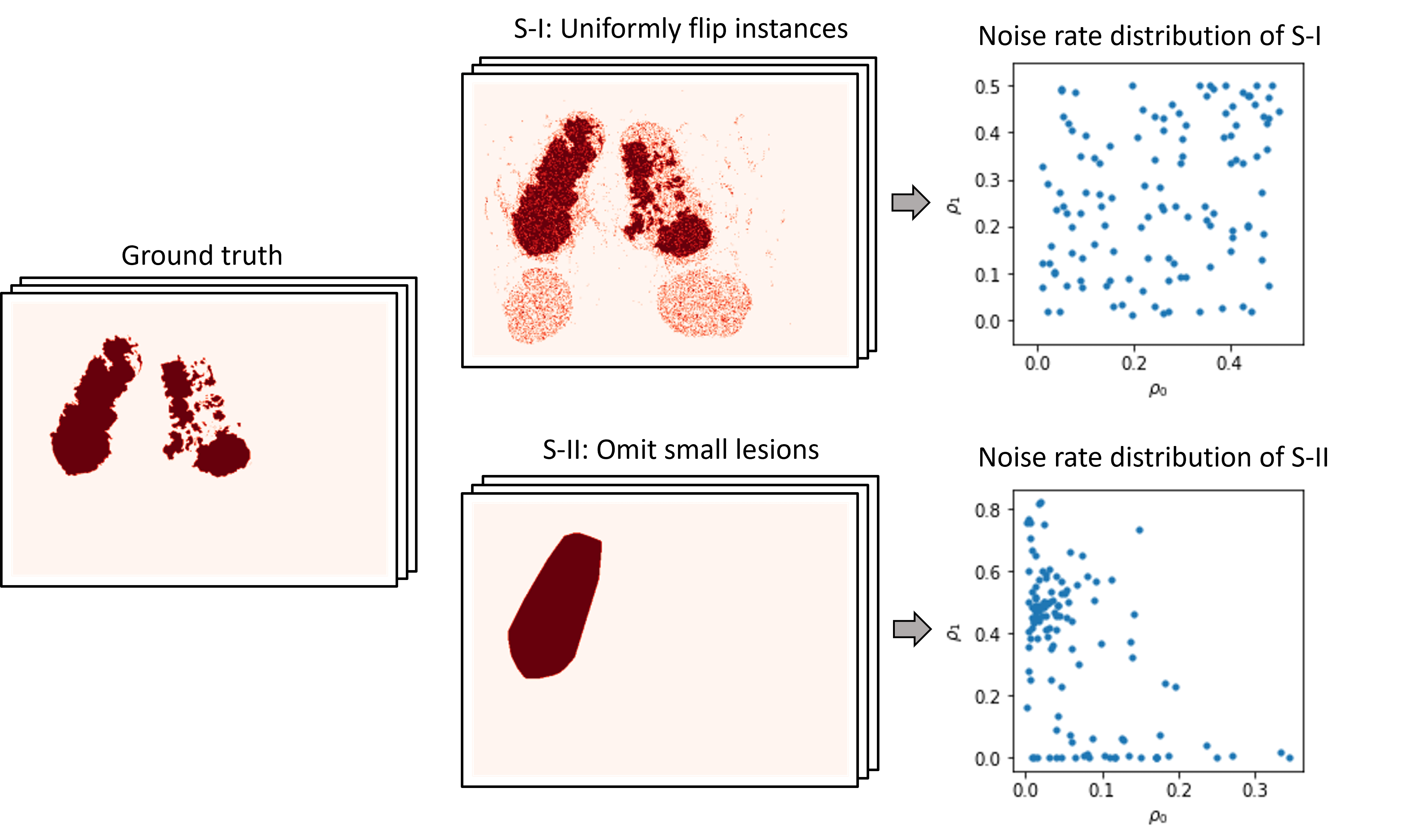}}
\caption{\textbf{Generation of synthetic coarse annotations} S-I: Coarse annotations generated by uniformly flipping patches; S-II: Coarse annotations generating by omitting small lesions. $\rho_0$: the fraction of mis-labeled instances in actual negative instances; $\rho_1$: the fraction of mis-labeled instances in actual positive instances.}
\label{fig:simulation-dataset}
\end{figure}

\begin{figure*}[t]
\centerline{\includegraphics[width=\textwidth]{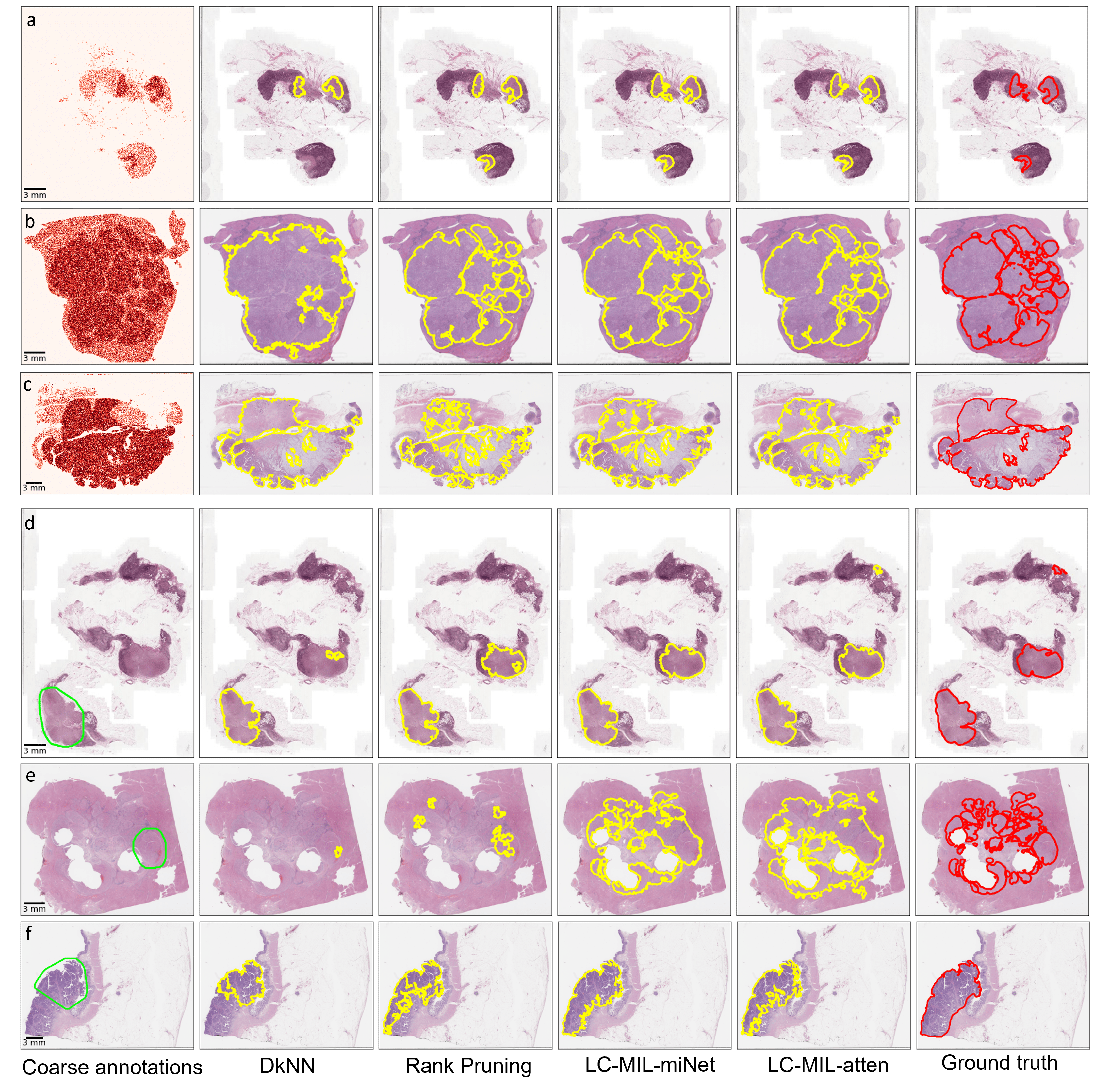}}
\caption{\textbf{Examples of synthetic coarse annotations and refinement.} (a-c) S-I: synthetic coarse annotations generated by uniformly flipping patches; (d-f): S-II: synthetic coarse annotations generated by omitting small lesions. Among those examples, (a) and (d) are from CAMELYON16, (b) and (e) are from PAIP2019, (c) and (f) are from PAIP2020. From left to right, six versions (coarse annotations (1st column, heatmap/lime lines); predicted contours using DkNN (2nd column, yellow lines); predicted contours using Rank Pruning (3rd column, yellow lines); predicted contours using LC-MIL-miNet (4th column, yellow lines)); predicted contours using LC-MIL-atten (5th column, yellow lines)); ground truth (6th column, red lines)) of cancerous regions contours are shown.}
\label{fig:examples-simulated}
\end{figure*}

\begin{table*}[]
\caption{ Summary of refinement on synthetic coarse annotations} 
\begin{tabular}{|c|ll|c|c|c|c|c|c|}
\hline
 &
  \multicolumn{2}{c|}{} &
  \multicolumn{3}{c|}{S-I: Uniformly Flip Instances} &
  \multicolumn{3}{c|}{S-II: Omit Small Lesions} \\ \cline{4-9} 
\multirow{-2}{*}{Datasets} &
  \multicolumn{2}{c|}{\multirow{-2}{*}{Methods}} &
  \multicolumn{1}{c|}{PPV} &
  \multicolumn{1}{c|}{TPR} &
  \multicolumn{1}{c|}{F1} &
  \multicolumn{1}{c|}{PPV} &
  \multicolumn{1}{c|}{TPR} &
  \multicolumn{1}{c|}{F1} \\ \hline
 &
  \multicolumn{2}{l|}{Coarse annotations} &
  \multicolumn{1}{c|}{0.545 ± 0.246} &
  \multicolumn{1}{c|}{0.733 ± 0.153} &
  \multicolumn{1}{c|}{0.592 ± 0.195} &
  \multicolumn{1}{c|}{0.573 ± 0.123} &
  \multicolumn{1}{c|}{0.606 ± 0.228} &
  \multicolumn{1}{c|}{0.561 ± 0.136} \\ 
 &
  \multicolumn{2}{l|}{DkNN} &
  \multicolumn{1}{c|}{0.798 ± 0.308} &
  \multicolumn{1}{c|}{0.649 ± 0.343} &
  \multicolumn{1}{c|}{0.686 ± 0.313} &
  \multicolumn{1}{c|}{0.810 ± 0.259} &
  \multicolumn{1}{c|}{0.523 ± 0.289} &
  \multicolumn{1}{c|}{0.597 ± 0.266} \\ 
 &
  \multicolumn{2}{l|}{Rank Pruning} &
  \multicolumn{1}{c|}{0.915 ± 0.050} &
  \multicolumn{1}{c|}{0.865 ± 0.148} &
  \multicolumn{1}{c|}{0.879 ± 0.113} &
  \multicolumn{1}{c|}{0.906 ± 0.083} &
  \multicolumn{1}{c|}{0.684 ± 0.204} &
  \multicolumn{1}{c|}{0.758 ± 0.144} \\ \cline{2-3} 
 &
  \multicolumn{1}{l|}{} &
  \multicolumn{1}{l|}{LC-MIL-miNet} &
  \multicolumn{1}{c|}{0.836 ± 0.151} &
  \multicolumn{1}{c|}{0.934 ± 0.062} &
  \multicolumn{1}{c|}{0.871 ± 0.104} &
  \multicolumn{1}{c|}{0.828 ± 0.120} &
  \multicolumn{1}{c|}{0.880 ± 0.132} &
  \multicolumn{1}{c|}{0.840 ± 0.091} \\ 
\multirow{-5}{*}{CAMELYON16} &
  \multicolumn{1}{l|}{\multirow{-2}{*}{LC-MIL}} &
  \multicolumn{1}{l|}{LC-MIL-atten} &
  \multicolumn{1}{c|}{0.863 ± 0.110} &
  \multicolumn{1}{c|}{0.915 ± 0.077} &
  \multicolumn{1}{c|}{\textbf{0.882 ± 0.068}} &
  \multicolumn{1}{c|}{0.833 ± 0.128} &
  \multicolumn{1}{c|}{0.895 ± 0.110} &
  \multicolumn{1}{c|}{\textbf{0.849 ± 0.084}} \\ \hline
 &
  \multicolumn{2}{l|}{Coarse annotations} &
  \multicolumn{1}{c|}{0.641 ± 0.237} &
  \multicolumn{1}{c|}{0.747 ± 0.151} &
  \multicolumn{1}{c|}{0.665 ± 0.182} &
  \multicolumn{1}{c|}{0.674 ± 0.123} &
  \multicolumn{1}{c|}{0.757 ± 0.266} &
  \multicolumn{1}{c|}{0.687 ± 0.168} \\ 
 &
  \multicolumn{2}{l|}{DkNN} &
  \multicolumn{1}{c|}{0.916 ± 0.149} &
  \multicolumn{1}{c|}{0.767 ± 0.317} &
  \multicolumn{1}{c|}{0.790 ± 0.291} &
  \multicolumn{1}{c|}{0.788 ± 0.185} &
  \multicolumn{1}{c|}{0.728 ± 0.303} &
  \multicolumn{1}{c|}{0.722 ± 0.242} \\ 
 &
  \multicolumn{2}{l|}{Rank Pruning} &
  \multicolumn{1}{c|}{0.911 ± 0.114} &
  \multicolumn{1}{c|}{0.915 ± 0.070} &
  \multicolumn{1}{c|}{0.907 ± 0.083} &
  \multicolumn{1}{c|}{0.858 ± 0.129} &
  \multicolumn{1}{c|}{0.755 ± 0.275} &
  \multicolumn{1}{c|}{0.768 ± 0.215} \\ \cline{2-3} 
 &
  \multicolumn{1}{l|}{} &
  \multicolumn{1}{l|}{LC-MIL-miNet} &
  \multicolumn{1}{c|}{0.875 ± 0.156} &
  \multicolumn{1}{c|}{0.942 ± 0.058} &
  \multicolumn{1}{c|}{0.875 ± 0.156} &
  \multicolumn{1}{c|}{0.828 ± 0.151} &
  \multicolumn{1}{c|}{0.889 ± 0.137} &
  \multicolumn{1}{c|}{\textbf{0.842 ± 0.119}} \\ 
\multirow{-5}{*}{PAIP2019} &
  \multicolumn{1}{l|}{\multirow{-2}{*}{LC-MIL}} &
  \multicolumn{1}{l|}{LC-MIL-atten} &
  \multicolumn{1}{c|}{0.895 ± 0.138} &
  \multicolumn{1}{c|}{0.939 ± 0.066} &
  \multicolumn{1}{c|}{\textbf{0.907 ± 0.098}} &
  \multicolumn{1}{c|}{0.884 ± 0.129} &
  \multicolumn{1}{c|}{0.826 ± 0.177} &
  \multicolumn{1}{c|}{0.834 ± 0.128} \\ \hline
 &
  \multicolumn{2}{l|}{Coarse annotations} &
  \multicolumn{1}{c|}{0.545 ± 0.246} &
  \multicolumn{1}{c|}{0.697 ± 0.153} &
  \multicolumn{1}{c|}{0.592 ± 0.195} &
  \multicolumn{1}{c|}{0.682 ± 0.098} &
  \multicolumn{1}{c|}{0.497 ± 0.079} &
  \multicolumn{1}{c|}{0.570 ± 0.076} \\ 
 &
  \multicolumn{2}{l|}{DkNN} &
  \multicolumn{1}{c|}{0.901 ± 0.220} &
  \multicolumn{1}{c|}{0.626 ± 0.387} &
  \multicolumn{1}{c|}{0.662 ± 0.372} &
  \multicolumn{1}{c|}{0.882 ± 0.086} &
  \multicolumn{1}{c|}{0.408 ± 0.113} &
  \multicolumn{1}{c|}{0.548 ± 0.118} \\ 
 &
  \multicolumn{2}{l|}{Rank Pruning} &
  \multicolumn{1}{c|}{0.900 ± 0.197} &
  \multicolumn{1}{c|}{0.807 ± 0.165} &
  \multicolumn{1}{c|}{\textbf{0.841 ± 0.176}} &
  \multicolumn{1}{c|}{0.910 ± 0.092} &
  \multicolumn{1}{c|}{0.477 ± 0.173} &
  \multicolumn{1}{c|}{0.607 ± 0.147} \\ \cline{2-3} 
 &
  \multicolumn{1}{l|}{} &
  \multicolumn{1}{l|}{LC-MIL-miNet} &
  \multicolumn{1}{c|}{0.820 ± 0.198} &
  \multicolumn{1}{c|}{0.836 ± 0.202} &
  \multicolumn{1}{c|}{0.811 ± 0.181} &
  \multicolumn{1}{c|}{0.874 ± 0.092} &
  \multicolumn{1}{c|}{0.736 ± 0.163} &
  \multicolumn{1}{c|}{\textbf{0.786 ± 0.106}} \\ 
\multirow{-5}{*}{PAIP2020} &
  \multicolumn{1}{l|}{\multirow{-2}{*}{LC-MIL}} &
  LC-MIL-atten &
  0.833 ± 0.191 &
  0.848 ± 0.175 &
  0.823 ± 0.167 &
  0.881 ± 0.108 &
  0.638 ± 0.192 &
  0.718 ± 0.138 \\ \hline
\end{tabular}
\label{tab:overview-synthetic}
\end{table*}

\subsection{Synthetic coarse annotations}
To fully evaluate the methods above, and before presenting our results on a real study with expert pathologists, we first study a synthetic setting where the experimental conditions (amounts of label corruptions, types of inaccurate annotations) can be easily varied. We generated coarse annotations automatically via two customs, detailed in the subsections that follow. For each WSIs, we considered the annotations provided by expert pathologists as the ground truth.

\subsubsection{S-I: Uniformly flipping patches}\label{sec: S-I}
A WSI with ground truth annotations provides a set of patches with correct labels. To generate simulated coarse annotations, we uniformly flip positive samples with noise rate $\rho_1$ (fraction of flipped positive samples) and negative samples with noise rate $\rho_0$ (fraction of flipped negative samples), allowing us to evaluate the performance of the different algorithms under a wide range of False Positives and False Negative rates. For each WSI, we generated a simulated coarse annotations in this way, where $\rho_1$ and $\rho_0$ are two independent random viable both uniformly distributed on $(0,0.5)$. 
Fig. \ref{fig:simulation-dataset} shows one example of the coarse annotations and the distribution of $\rho_1$ and $\rho_0$ on the whole dataset.

\subsubsection{S-II: Omitting small lesions}
To better simulate a clinically-relevant context, we also generate noisy annotations that are similar to those provided by human annotators (note that real expert annotations will also be addressed shortly). To this end, we create coarse annotations by the following three steps: (1) Retaining only the largest cancerous region and omit all the others; (2) Performing the morphological operation of dilation to the remaining cancerous region; (3) Taking the convex hull for the dilated lesion. As a particular case, since almost all of the WSIs in PAIP2020 contain only one cancerous region, the only cancerous region was cut in half. We also give one example of the coarse annotations and noise rate distribution on the whole dataset, as shown in Fig. \ref{fig:simulation-dataset}.

These two procedures described above had the purpose of simulating both false positives as well as false negatives in the initial inaccurate annotation. The proposed method (LC-MIL) and all other comparative methods were applied to each WSI independently to refine the coarse annotations. Fig. \ref{fig:examples-simulated} shows a few examples of the refinement performance, and the reader can find more examples in Fig. \ref{fig:examples-simulated-supp}.

In order to quantitatively evaluate how inaccurate the coarse annotations are from the actual precise annotations, as well as the improvement obtained after applying our refinement method, we calculate precision (PPV), recall (TPR), and F1 scores for annotations before and after refinement. The results are summarized in Table \ref{tab:overview-synthetic}. Note that all the metrics were calculated per slide, and the average and standard deviation in each subset are reported. The reader can find other evaluation metrics, including specificity (TNR), negative predictive value (NPV) and intersection over Union (IoU) in Table \ref{tab:overview-S-I-all} and Table \ref{tab:overview-S-II-all}. To explore how different methods perform under different noise conditions, we sort the coarse annotations according to their F1 score, and then plot F1 scores of refined annotations, as shown in Fig. \ref{fig:scatterplot-simulated}. 

\textbf{Performance comparison}: The proposed method, LC-MIL, significantly improves disease localization of coarse annotations and corrects incorrect labels, and in general outperforms the comparative methods based on the F1 scores. DkNN typically achieves lower metrics than Rank Pruning and LC-MIL, and only bring slight improvement for the original inaccurate annotations, which illustrates the difficulty of the learning problem under consideration. Rank Pruning generally shows stronger improvement capacity compared with DkNN, while still slightly under-performing the proposed method. Moreover, LC-MIL is especially useful in detecting missed lesions. If we focus on two columns (\textit{i.e.}, PPV and TPR) in Table \ref{tab:overview-synthetic}, we will find that all the methods can efficiently improve PPV (\textit{i.e.}, making tumor boundaries more precise), but LC-MIL shows obvious advantages in improving TPR (\textit{i.e.}, detecting missed lesions). On the other hand, Rank Pruning generally performs best in improving PPV, and generate less false positives.


\textbf{Run-time comparison:} All the methods are implemented in Pytorch and trained on a single NVIDIA GTX1080Ti  GPU. All the methods make predictions on the same number of patches for each WSI. In general, LC-MIL is the fastest method, compared with DkNN, which needs to conduct k-NN computation on extremely large matrix, and Rank Pruning, which needs repeat training in a cross-validation manner. Note that the run-time merely depends on the number of patches of each WSI, and is not correlated with the the noise condition of coarse annotations. We randomly select 5 slides from each subset, and run each method in exactly the same setting in order to report the run-time comparison summarized in Table \ref{tab:runtime}.

\begin{table}[]
\centering
\caption{Run-time comparison (min).}
\begin{tabular}{|l|c|c|c|}
\hline
\multicolumn{1}{|l|}{Methods}   & CAMELYON16 & PAIP2019   & PAIP2020   \\ \hline
\multicolumn{1}{|l|}{DkNN}         & 6.94 ± 3.07  & 38.9 ± 18.3 & 15.8 ± 3.60 \\ \hline
\multicolumn{1}{|l|}{Rank Pruning} & 11.3 ± 2.69 & 34.9 ± 18.1 & 17.8 ± 3.90 \\ \hline
\multicolumn{1}{|l|}{LC-MIL-miNet}  & 5.60 ± 1.52 & 18.2 ± 5.95 & 9.84 ± 1.65  \\ \hline
\multicolumn{1}{|l|}{LC-MIL-atten}  & 5.70 ± 1.48 & 18.8 ± 5.87 & 11.8 ± 1.94 \\ 
\hline
\end{tabular}
\label{tab:runtime}
\end{table}

\begin{table*}[]
\centering
\caption{ Summary of refinement on hand-drawn coarse annotations} 
\begin{tabular}{|c|ll|c |c|l|}
\hline
Dataset &
  \multicolumn{2}{c|}{Method} &
  PPV &
  TPR &
  \multicolumn{1}{c|}{F1} \\ \hline
 & \multicolumn{2}{l|}{Coarse annotations} & 0.656 ±0.219  & 0.830 ±0.208  & 0.682 ± 0.149           \\
 & \multicolumn{2}{l|}{DkNN}               & 0.832 ±0.167  & 0.776 ±0.252  & 0.756 ± 0.182           \\ 
 & \multicolumn{2}{l|}{Rank Pruning}       & 0.868 ±0.131  & 0.836 ±0.168  & 0.830 ± 0.121           \\ \cline{2-3} 
 & \multicolumn{1}{l|}{}    & LC-MIL-miNet    & 0.819 ±0.141  & 0.898 ±0.129  & 0.840 ± 0.096           \\ 
\multirow{-5}{*}{CAMELYON16} &
  \multicolumn{1}{l|}{\multirow{-2}{*}{LC-MIL}} &
  LC-MIL-atten &
  0.828 ± 0.140 &
  0.901 ± 0.114 &
  \textbf{0.848 ± 0.096} \\ \hline
 & \multicolumn{2}{l|}{Coarse annotations} & 0.647 ±0.202  & 0.934 ±0.174  & 0.742 ± 0.182           \\
 & \multicolumn{2}{l|}{DkNN}               & 0.753 ±0.202  & 0.920 ±0.179  & 0.808 ± 0.181           \\  
 & \multicolumn{2}{l|}{Rank Pruning}       & 0.799 ±0.194  & 0.913 ±0.155  & 0.830 ± 0.168  \\ \cline{2-3} 
 & \multicolumn{1}{l|}{}    & LC-MIL-miNet    & 0.785 ±0.202  & 0.932 ±0.131  & \textbf{0.833 ± 0.167}           \\ 
\multirow{-5}{*}{PAIP2019} &
  \multicolumn{1}{l|}{\multirow{-2}{*}{LC-MIL}} &
  LC-MIL-atten &
  0.805 ± 0.213 &
  0.919 ± 0.114 &
  \textbf{0.833 ± 0.163} \\ \hline
 & \multicolumn{2}{l|}{Coarse annotations} & 0.705 ± 0.152 & 0.951 ± 0.092 & 0.796 ± 0.114          \\ 
 & \multicolumn{2}{l|}{DkNN}               & 0.882 ± 0.105 & 0.936 ± 0.112                         & \textbf{0.898 ± 0.095} \\ 
 & \multicolumn{2}{l|}{Rank Pruning}       & 0.908 ± 0.091 & 0.853 ± 0.152 & 0.867 ± 0.110        \\ \cline{2-3} 
 & \multicolumn{1}{l|}{}    & LC-MIL-miNet    & 0.860 ± 0.112 & 0.897 ± 0.103 & 0.869 ± 0.081          \\ \
\multirow{-5}{*}{PAIP2020} &
  \multicolumn{1}{l|}{\multirow{-2}{*}{LC-MIL}} &
  LC-MIL-atten &
  0.902 ± 0.097 &
  0.844 ± 0.114 &
  0.863 ± 0.077 \\ \hline
\end{tabular}
\label{tab:overview-real-world}
\end{table*}

\begin{figure*}[t]
\centerline{\includegraphics[width=\textwidth]{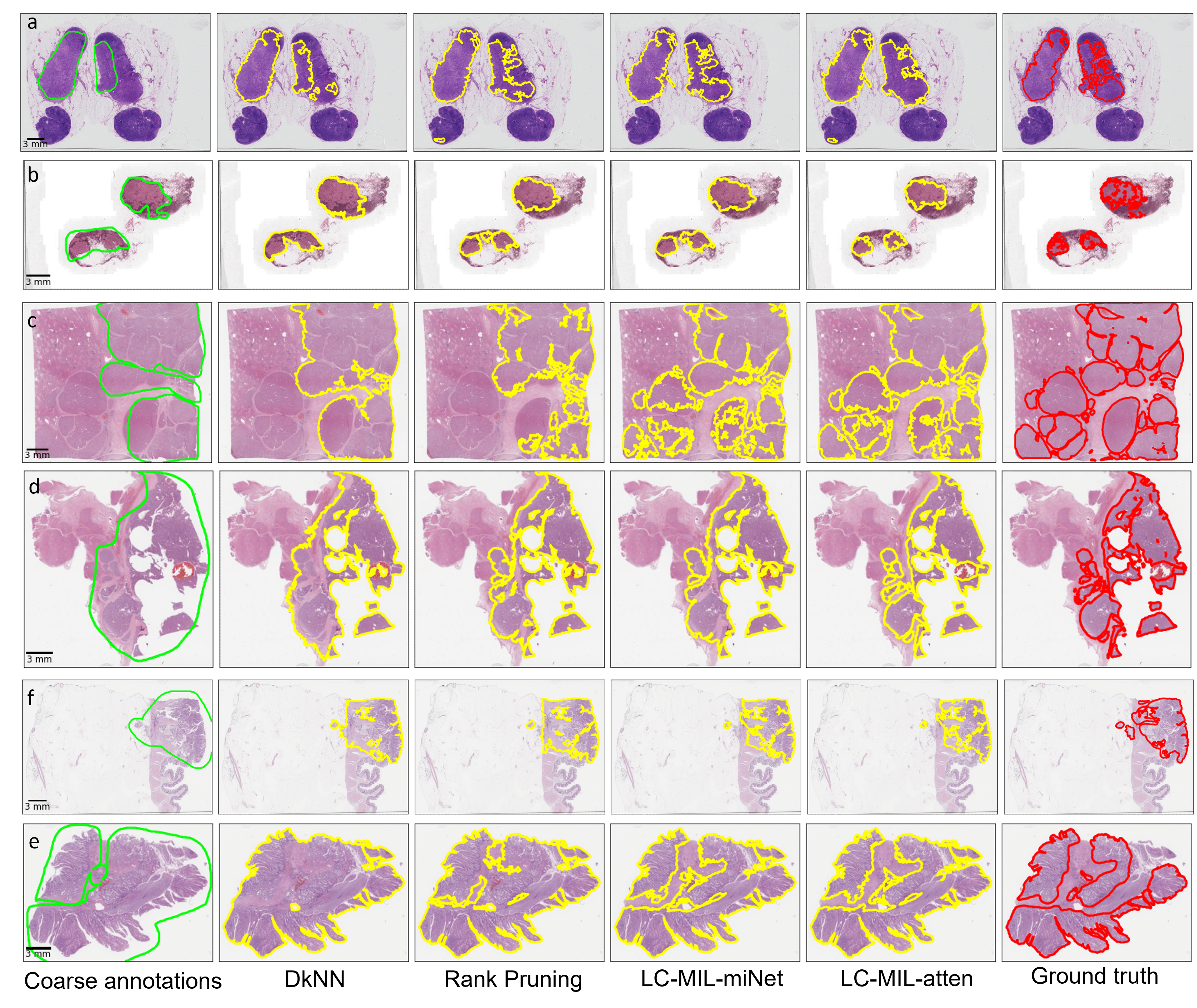}}
\caption{\textbf{Examples of hand-drawn coarse annotations and refinement.} Among those examples, (a) and (b) are from CAMELYON16, (c) and (d) are from PAIP2019, (e) and (f) are from PAIP2020. From left to right, six versions (coarse annotations (1st column, lime lines); predicted contours using DkNN (2nd column, yellow lines); predicted contours using Rank Pruning (3rd column, yellow lines); predicted contours using LC-MIL-miNet (4th column, yellow lines)); predicted contours using LC-MIL-atten (5th column, yellow lines)); ground truth (6th column, red lines)) of cancerous regions contours are shown.}
\label{fig:examples-real-world}
\end{figure*}

\subsection{Real-world experiment: improving pathologists' annotations}

To evaluate the proposed method in a real-world scenario, we collaborated with two expert (senior) pathology residents from Johns Hopkins Medicine. Since the datasets we employ already have ground truth annotations provided by pathologists, we explore whether our proposed formulation can be used to improve the pathologists' workflow by refining approximate segmentation. In particular, pathologists were asked to provide a segmentation of the tumor regions in each of the 120 WSIs, but to do so in a time constrained manner, with 30 seconds per slide. We regard these as the coarse annotations to refine. 

The proposed LC-MIL method, as well as other comparative methods, were applied in order to refine these quick and coarse annotations. Note that the refinement was still conducted on each slide and each pathologist's annotations independently, and we did not use any methods to incorporate annotations from different pathologists. Fig. \ref{fig:examples-real-world} shows some examples, and the overall refinement results are summarized in Table \ref{tab:overview-real-world}. The reader can also find more evaluation metrics in Table \ref{tab:overview-real-world-all} and more examples in Fig. \ref{fig:examples-real-world-supp}. We also show the distributions of coarse annotations drawn by pathologists in a time-constrained manner, as well as the refinement performance in different noise conditions, as Fig. \ref{fig:scatterplot-real-world} shows.

As Table \ref{tab:overview-real-world} shows, all methods improves the coarse annotations to some extend on all three datasets. LC-MIL achieves the best overall performance, obtaining the second-best F1 only on PAIP2020. Interestingly, the best method for refining coarse annotations on colorectal samples (PAIP2020) is DkNN, which achieved the lowest F1 score in the synthetic datasets. We believe that this is mainly caused by the heterogeneity and complexity of colon histology: Colon tissue samples typically contain loose connective tissue, smooth muscles, and epithelial tissue, where the cancer begins developing. The texture difference between benign and malignant epithelial tissues is much more subtle than the dissimilarity across different tissue types. The LC-MIL model seems to be more easily misled to discriminate epithelial tissues from any other regions, resulting in some normal epithelial tissues being detected as (false) positive. On the other hand, when pathologists annotate colorectal cancer, they make use of both local (e.g., texture) and global (e.g., location, shape) information, and thus rarely mis-classify normal epithelial tissues. 
It is worth noting that pathologists' annotations are sufficiently good on those samples, rarely missing lesions (i.e. high TPR), and the only imperfection resides on rough boundaries. DkNN seems to be particularly suitable for this condition, as it might be the most conservative among all implemented method --keeping the overall segmentation while make only local improvement on boundaries-- and explaining the difference in observed performance. More broadly, handling tissue samples with confounding morphology can a challenging problem, and we expanded this discussion in the Section \ref{sec:Discussion}.

\section{Implementation details and sensitivity to parameters}\label{sec:implementation-details}
In addition to the central machine learning component of our method, detailed above in Section \ref{sec:Methods}, these experiments involve other implementations details common in image analysis pipelines, which we detail next.

\begin{figure}[h]
\centerline{\includegraphics[width=\columnwidth]{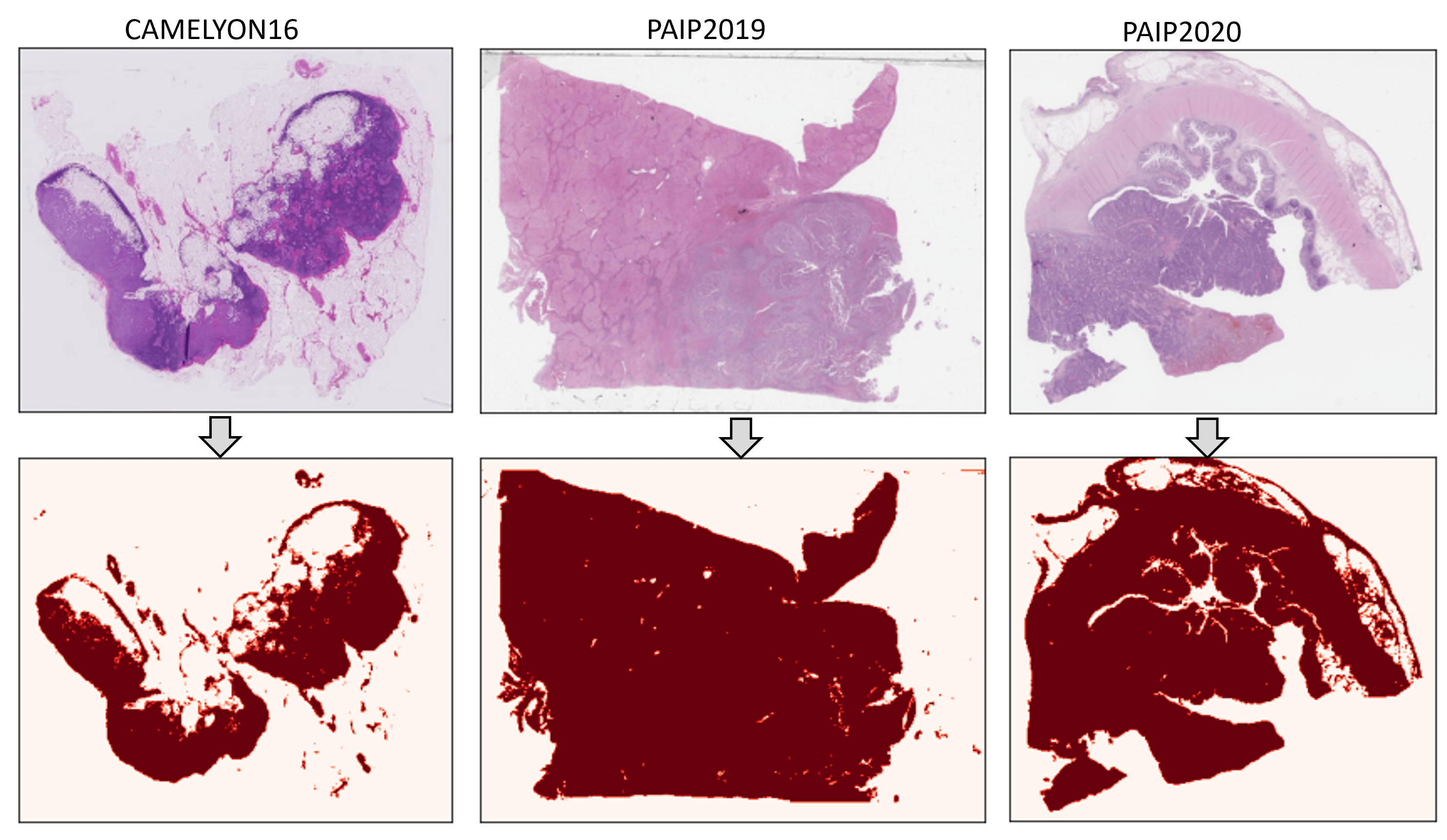}}
\caption{\textbf{Examples of tissue region identification.} One example of each subset is shown. The first rows shows thumbnails of WSIs, and the second row shows the corresponding tissue masks.}
\label{fig:pre-processing}
\end{figure}

\textbf{Tissue region identification}: For each digitized slide, our pipeline begins with the automatic detection of tissue regions to exclude irrelevant (e.g. blank) sections. Gigapixel WSIs are first loaded into memory at a down-sampled resolution (\textit{e.g.}, $256\times$ downscale), which we use for the detection of empty regions. The detection methods vary slightly for different data source. For CAMELYON16, the downsampled thumbnail is converted from RGB color space to the HSV color space. Otsu's algorithm \cite{4310076} is applied to the H and S channels independently and then two masks are combined (employing logic \textsc{and}) to generate the final binary mask. For PAIP2019 and PAIP2020, the tissue mask is generated by applying the RGB thresholds $(235, 210, 235)$ on images, as suggested by the data provider. We provided some examples of the tissue masks in Fig. \ref{fig:pre-processing}. 

\textbf{WSI patching}: The size of each patch is set to $256\times256$ (pixels). Technically, every pixel can be the center of a unique patch, resulting in millions of patches that highly overlap with each other. In this way, the predicted heatmaps or lesion maps have the same size and resolution as the original WSIs. However, such an approach is extremely computationally intensive and time consuming, turning it prohibitive. In our experiments, we extracted patches with no overlap in WSIs scanned at \(\times\)40 magnification (CAMELYON16 and PAIP2020). For slides scanned at \(\times\)20 magnification (PAIP2019), cropped patches have 75\% overlap with neighborhood patches. The size and overlap of cropped patches naturally impact the resolution of predicted heatmaps. For CAMELYON16 and PAIP2020, the predicted heatmap is $256\times$ downscaled of the original WSI; for PAIP2019, the predicted heatmap is $128\times$ downscaled of the original WSI. Our overall approach is not limited to these choices, and could be applicable to other settings too.

\textbf{Neural networks architecture}: The feature extractors used in LC-MIL (\(f_{\psi}\)) are built based on the 16-layer VGGNet architecture \cite{Simonyan2015}. The last FCN layer is removed to generate \(f_{\psi}\). Since the training data is limited (one WSI), the VGGNet used has been pre-trained on the ImageNet dataset \cite{Deng2010}. When they are further fine-tuned on the histopathology images, the parameters in the convolutional layers are kept constant.

\textbf{Learning hyper-parameters:} The MIL model used in LC-MIL is trained on 1,000 MIL bags, consisting of 500 positive bags and 500 negative bags. Adam optimizer is used with an initial learning rate of $5\times10^{-5}$, and the learning rate decays to 50\% for every 100 bags. For CAMELYON16 and PAIP2019, each MIL bag contains 10 instances; for PAIP2020, each bag contains 3 instances. We discuss the sensitivity to the choice of number of bags and bag size below.

\textbf{Post-processing}: To obtain a fair comparison between the annotations before and after refinement, we conduct some simple post-processing on the heatmaps generated by either DkNN (binary map), Rank Pruning (scalar map), and LC-MIL (scalar map). The binary heatmap predicted by DkNN is post-processed by simple morphology operations: both small holes (smaller than 100 pixels) and small objects (smaller than 100 pixels) are removed. The scalar heatmap predicted by Rank Pruning and LC-MIL is first converted to a binary map using thresholding. The threshold $v_0$ is decided by applying Otsu's algorithm \cite{4310076} to the predicted scores of instances that are originally annotated positive, denoted as $v_0 = \text{OTSU}(\{p_i| x_i \in S_p\})$. Finally we applied the same morphology operations conducted for DkNN to these binary maps. 


\subsection{Sensitivity to parameters}\label{sec:para-sens}
\subsubsection{Using multiple slides}\label{sec:multiple-slides}
\begin{figure}[h]
\centerline{\includegraphics[width=\columnwidth]{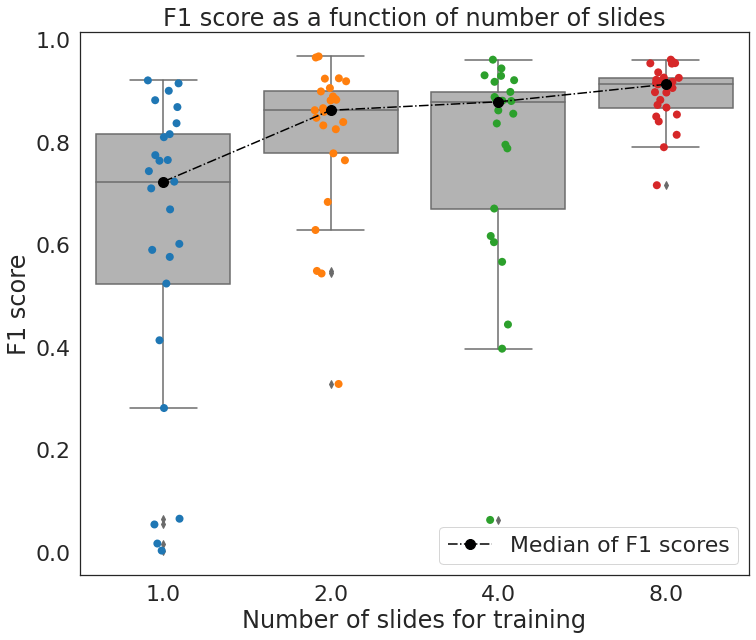}}
\caption{Prediction results using multiple slides for training.}
\label{fig:para-sensi-num-slides}
\end{figure}

A natural extension of the proposed method is to incorporate multiple slides for training when a larger dataset is available. This can be implemented by slightly modifying the LC-MIL framework described in Section \ref{sec:method-LC-MIL}. To be more specific, MIL bags can be constructed from each single slide in exactly the same way, as described in Eq. \ref{eq:MIL set}. After that, these MIL bags from different slides can be combined to form a larger MIL dataset, denoted as $S_\text{MIL}^{\prime} = \{ S_\text{MIL}^{(0)}, ..., S_\text{MIL}^{(k) }\}$, where $k$ is the number of slides. The MIL model is then trained on this larger MIL dataset $S_\text{MIL}^{\prime}$. At inference stage, the trained MIL model is used to predict tumor segmentation on held-out slides. \\
We empirically tested this new implementation and explored how the number of slides affect the prediction performance. The detailed setting is described as follows. We randomly split the 54 WSIs in PAIP2019 into training set (49 slides), and held-out validation set (5 slides). The MIL model is trained on $k$ slides (with coarse annotations) that are randomly selected from the training set, and tested on the held-out validation set. We set $k = 1, 2, 4, 8$ to explore different conditions. At each $k$, the training-validation process is repeated 5 times to alleviate the impacts of extreme cases. We plot the F1 scores of the prediction results on the held-on validation set in Fig. \ref{fig:para-sensi-num-slides}. As Fig. \ref{fig:para-sensi-num-slides} shows, as $k$ increases, the obtained F1 scores of the predicted annotations increase, and their variance decreases. This experiment demonstrates how the proposed LC-MIL framework could be adjusted in a multi-slides setting and make predictions on held-out sets containing slides with no annotations. Naturally, this experiment also demonstrates that larger training set leads to a better and more stable system. 
\subsubsection{Choice of bag size and bag numbers}
\begin{figure}[h]
\centerline{\includegraphics[width=0.8\columnwidth]{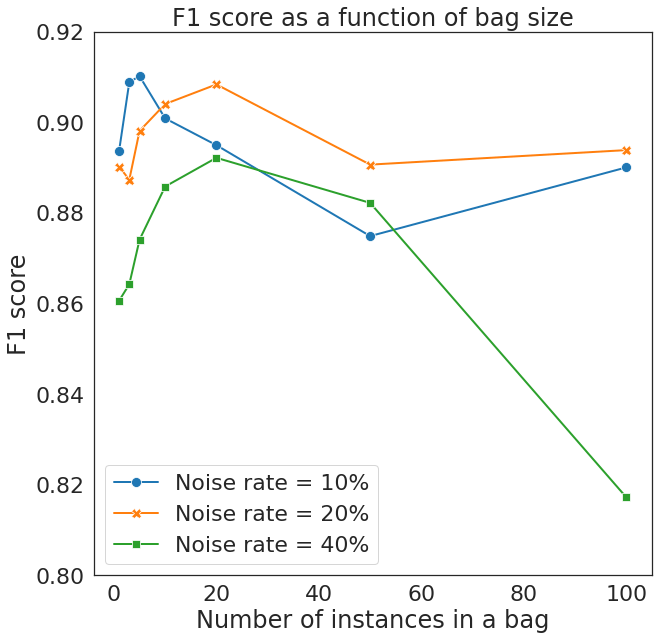}}
\caption{Prediction results using various bag size.}
\label{fig:para-sensi-bag-size}
\end{figure}

As presented in Eq. \ref{eq:MIL set}, there are two adjustable parameters in the construction of MIL sets: the number of instances within a bag (i.e. bag size), denoted as $n_j$, and the number of bags, denoted as $M$. Here we give the detailed settings for those parameters, and discuss the sensitivity to the their respective choices.
\textbf{Bag size $n_j$:} The bag size is affected by the amount of label noise as well as the complexity of the task (in terms of the heterogeneity of the histopathology images), and both aspects should be considered in determining an optimal bag size. We first conduct an experiment to discuss of the choice of bags size in different noise conditions. We randomly select 4 slides, and generate three versions of coarse annotations for each of them. To explicitly control the noise rates, those coarse annotations are generated by uniformly flipping patches, as we described in Section \ref{sec: S-I}. The noise rates for these three versions of coarse annotations are: (1) $\rho_0 = \rho_1 = 10\%$; (2) $\rho_0 = \rho_1 = 20\%$; (3) $\rho_0 = \rho_1 = 40\%$. For each slide with a specific version of coarse annotations, we applied LC-MIL-atten to refine the annotations, and the mean F1 scores of refined annotations are summarized in Fig.\ref{fig:para-sensi-bag-size}. 
As shown in Fig. \ref{fig:para-sensi-bag-size}, we suggest that a bag size around 10 can be considered as an appropriate setting in general. When noise rate is low (e.g., $\rho_0 = \rho_1 = 10\%$), a smaller bag size achieves the best result. This is consistent with intuition -- in the extreme case where coarse annotations are exactly the ground truth (i.e. no label noise exists), a bag size of 1 should be the best choice. Naturally, this case reduces the MIL classifier to a standard instance-level binary classifier. On the other hand, we do not suggest a very large bag size (e.g., $n_j > 50$) in all circumstances. Although a large bag size guarantees that the majority of instances within a MIL bag are correctly labeled, it also bring substantial ambiguity of instance-level prediction, and might decrease the F1 scores of patch-wise classification. 

Furthermore, we recommend to use a smaller bag size for tissues with compound morphology, such as the colorectal tissues. In the colorectal slides, normal epithelial tissues shows similar textures with the cancerous tissues, while other components (e.g., fat tissue, smooth muscles) have significantly different textures, simply because they are different types of tissues. When human identify tumor in colorectal tissues, both local texture and global information are taken into account. On the other hand, LC-MIL can purely utilize local texture information, and tend to mis-classify normal epithelial tissues as tumor, especially when the bag size is large. Utilizing a smaller bag size in this case leads to a more conservative refinement system, and thus decreases false negatives. 

In this work, we used $n_j =  10$ for slides in CAMELYON16 (lymph node metastasis of breast cancer) and PAIP2019 (liver cancer), and $n_j = 3$ for slides in PAIP2020 (colorectal cancer) considering its complex and heterogeneous morphology. Note however that these parameters were not optimized per case. Further improvements are to be expected if these parameters are adjusted based on the label noise conditions and specifics histology in the WSI. We leave this as a future direction of the proposed method.   

\begin{figure}[h]
\centerline{\includegraphics[width=0.8\columnwidth]{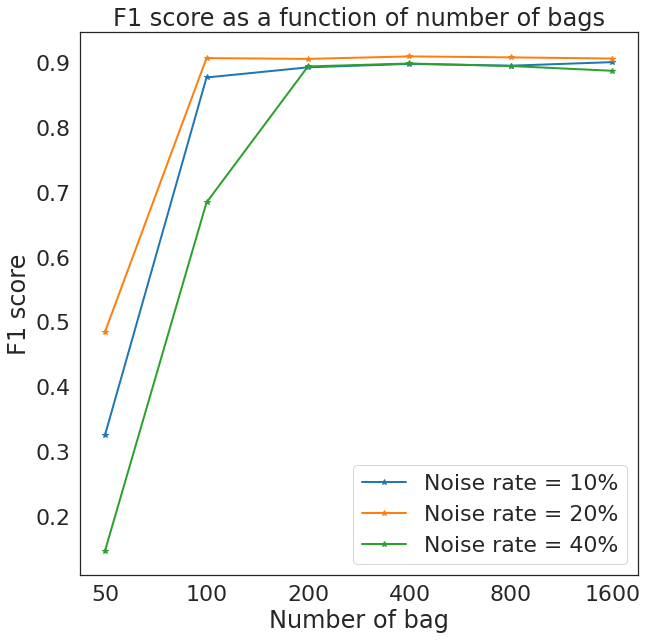}}
\caption{Prediction results using various number of bags.}
\label{fig:para-sensi-bag-number}
\end{figure}

\textbf{Bag number $M$:} The number of bags represent the number of training samples provided to the multiple instance learner. Consequently, a minimal number of $M$ is expected for the algorithm to achieve a reasonable performance. We will now demonstrate the this number is relatively small, so that in all cases the models are able to have sufficient training data. To explore the effect of bag number $M$, we use a constant bag size of 10, and gradually increase $M$ to see how results change. We used the same slides and coarse annotations as those used in discussion of bag size $n_j$. The F1 scores of refined annotations are summarized in Fig. \ref{fig:para-sensi-bag-number}, demonstrating that 400 bags suffice to obtain reasonable results in the single-slide refinement setting. In this work, we set $M = 1000$ for each single slide in all experiments. 

\section{Discussion}\label{sec:Discussion}
The main hypothesis studied in this work is that the coarse annotations on whole-slide images can be refined automatically even from very limited data. This problem had remained unstudied until now due to the technical challenges that often make deep learning models require very large training sets, and could alleviate the workload of expert pathologists.

To test our hypothesis, we developed the Label Cleaning Multiple Instance Learning (LC-MIL) method to refine the coarse annotations. Our experiments on a heterogeneous dataset with 120 WSIs across three different types of cancers show that LC-MIL can be used to generate a significantly better version of disease segmentation, even while learning from a single WSI with very coarse annotations. When compared to other baseline alternatives specifically adapted to this setting (since again, this problem has not been studied before), our proposed algorithm generally performs the best across all cases considered. 

Our methodology has implications in both engineering and clinical domains. From a machine learning perspective, LC-MIL demonstrates the potential of the MIL framework to be used in a label cleaning context. Our setting can be considered as a significant complement to existing label cleaning methods, without requiring the need for an auxiliary clean set of samples. From a clinical perspective, the proposed methodology could be used to alleviate pathologists' workload in annotation tasks. LC-MIL allows pathologists to draw coarse annotations quickly and obtain a refined version for reference. On the other hand, our results show that the refinement produced by LC-MIL is especially significant when the coarse annotations are very inaccurate, which implies that the proposed methodology could be useful for helping those very inexperienced annotators. Importantly, our approach is particularly useful for increasing TPR, therefore detecting tumor areas that might have been missed by an inexperienced, or distracted, annotator.

A significant advantage of the LC-MIL is its ability to work in extreme data-scarcity scenarios, as it can be deployed on a single WSI. We have chosen to study this setting to showcase the flexibility of our approach. However, this also constrains the minimal and maximal ratio of cancerous regions if the refinement is conducted per slide, since a minimal amount of both negative and positive samples within one slide are needed for the learning algorithm. This limitation could be addressed by moving beyond the \emph{single} WSI case and  aggregating patches from multiple slides during the learning phase, if such data is available. Moving from this single-slide to multiple-slides settings is a natural extension of our methodology, as we demonstrated in Section \ref{sec:multiple-slides}.

Another limitation of LC-MIL, probably also shared by other binary classifiers, is that the model can be confused by irrelevant information when the histology structure is complex and heterogeneous. An example of this is that the LC-MIL model tends to be misled to discriminate epithelial tissue from other types of tissues (loose connective tissue, smooth muscle), instead of focusing on learning the difference between benign and malignant tissue. On the other hand, this may not be a challenging task for the human (and experienced) eye. We suggest that this could be addressed by extending the current version of LC-MIL to a multi-class setting. Moreover, pathologists usually make use of both local texture information and global information (e.g., shape, location) to make decisions, while the proposed LC-MIL, as well as other patch-based method, merely depend on local texture and fail to incorporate global information. Efficiently incorporating this global context can be a challenging and significant problem, which we also leave as future work.

An important open question remains: can detailed and labor-intensive annotations be replaced by simple and quick coarse annotations, with the help of our methodology? In our real-world experiment, where expert pathologists made quick annotations that were later refined by our LC-MIL, about 0.85 F1 scores were obtained (from initial values of about 0.73, an improvement of 16\%), having taken only 30 seconds for each slide. Moreover, it remains unclear what the inter- and intra-reader variability of these annotations are. Judging by other studies in different contexts \cite{Foss2012, Loes2013}, it is natural to assume an F1 scores of 1 is unrealistic across readers, and that F1 scores in the range of 0.9-1 might suffice. Further and larger scale studies should be designed to thoroughly evaluate this possibility.

\section{Conclusion}\label{sec:Conclusion}
Altogether, in this paper we developed LC-MIL, a label cleaning method under a multiple instance learning framework, to automatically refine coarse annotations on a single WSI. The proposed methodology demonstrates the potential of the MIL framework in the label cleaning context, and provides a new way of deploying MIL models so that they can trained on even a single slide. LC-MIL holds promise to relieve the workload of pathologists, as well as in helping inexperienced annotators in challenging annotation tasks.

\section*{Acknowledgment}
The authors thank Dr. Ashley M. Cimino-Mathews and Dr. Marissa Janine White from the Department of Pathology, Johns Hopkins University, School of Medicine, for their valuable advice and insights in histopathology. The authors also kindly thank Dr. Adam Charles, Haoyang Mi and Jacopo Teneggi from the Department of Biomedical Engineering, Johns Hopkins University, for their useful advice and discussions.

\bibliographystyle{IEEEtran}
\bibliography{main}

\clearpage
\newpage
\setcounter{figure}{0}
\setcounter{table}{0}
\renewcommand\thefigure{A. \arabic{figure}}
\renewcommand\thetable{A. \arabic{table}}


\begin{table*}[!h]

\centering
\caption{Summary of refinement on S-I: Uniformly flip patches}
\begin{tabular}{|c|c|ll|c|l|c|c|c|l|}
\hline
Dataset &
  Tumor/tissue  &
  \multicolumn{2}{c|}{Method} &
  F1 &
  \multicolumn{1}{c|}{IoU} &
  PPV &
  TPR &
  TNR &
  \multicolumn{1}{c|}{NPV} \\ \hline
 &
  &
  \multicolumn{2}{l|}{Coarse annotations} &
  0.592 ± 0.195 &
  0.416 ± 0.189 &
  0.545 ± 0.246 &
  0.733 ± 0.153 &
  0.745 ± 0.137 &
  0.869 ± 0.139 \\ 
 &
  &
  \multicolumn{2}{l|}{DkNN} &
  0.686 ± 0.313 &
  0.570 ± 0.297 &
  0.798 ± 0.308 &
  0.649 ± 0.343 &
  0.971 ± 0.038 &
  0.889 ± 0.146 \\ 
 &
  &
  \multicolumn{2}{l|}{Rank Pruning} &
  0.879 ± 0.113 &
  0.772 ± 0.147 &
  0.915 ± 0.050 &
  0.865 ± 0.148 &
  0.965 ± 0.033 &
  0.956 ± 0.040 \\ \cline{3-4} 
 &
  &
  \multicolumn{1}{l|}{} &
  LC-MIL-miNet &
  0.871 ± 0.104 &
  0.758 ± 0.144 &
  0.836 ± 0.151 &
  0.934 ± 0.062 &
  0.912 ± 0.147 &
  0.971 ± 0.037 \\ 
\multirow{-5}{*}{\shortstack{CAME \\ LYON16  }} &
  \multirow{-5}{*}{\shortstack{29.7 \% \\(± 17.0 \%)}} &
  \multicolumn{1}{l|}{\multirow{-2}{*}{LC-MIL}} &
  LC-MIL-atten &
  \textbf{0.882 ± 0.068} &
  0.770 ± 0.117 &
  0.863 ± 0.110 &
  0.915 ± 0.077 &
  0.938 ± 0.059 &
  0.967 ± 0.042 \\ \hline
 &
  &
  \multicolumn{2}{l|}{Coarse annotations} &
  0.665 ± 0.182 &
  0.514 ± 0.197 &
  0.641 ± 0.237 &
  0.747 ± 0.151 &
  0.758 ± 0.157 &
  0.819 ± 0.144 \\ 
 &
  &
  \multicolumn{2}{l|}{DkNN} &
  0.790 ± 0.291 &
  0.711 ± 0.297 &
  0.916 ± 0.149 &
  0.767 ± 0.317 &
  0.961 ± 0.057 &
  0.905 ± 0.140 \\ 
 &
  &
  \multicolumn{2}{l|}{Rank Pruning} &
  0.907 ± 0.083 &
  0.823 ± 0.121 &
  0.911 ± 0.114 &
  0.915 ± 0.070 &
  0.952 ± 0.056 &
  0.950 ± 0.039 \\ \cline{3-4} 
 &
  &
  \multicolumn{1}{l|}{} &
  LC-MIL-miNet &
  0.875 ± 0.156 &
  0.811 ± 0.144 &
  0.875 ± 0.156 &
  0.942 ± 0.058 &
  0.919 ± 0.164 &
  0.954 ± 0.054 \\ 
\multirow{-5}{*}{PAIP2019} &
  \multirow{-5}{*}{\shortstack{38.0 \% \\(± 18.9 \%)}} &
  \multicolumn{1}{l|}{\multirow{-2}{*}{LC-MIL}} &
  LC-MIL-atten &
  \textbf{0.907 ± 0.098} &
  0.828 ± 0.133 &
  0.895 ± 0.138 &
  0.939 ± 0.066 &
  0.934 ± 0.130 &
  0.957 ± 0.046 \\ \hline
 &
  &
  \multicolumn{2}{l|}{Coarse annotations} &
  0.592 ± 0.195 &
  0.503 ± 0.169 &
  0.545 ± 0.246 &
  0.697 ± 0.153 &
  0.750 ± 0.138 &
  0.765 ± 0.157 \\ 
 &
  &
  \multicolumn{2}{l|}{DkNN} &
  0.662 ± 0.372 &
  0.588 ± 0.362 &
  0.901 ± 0.220 &
  0.626 ± 0.387 &
  0.965 ± 0.058 &
  0.830 ± 0.168 \\ 
 &
  &
  \multicolumn{2}{l|}{Rank Pruning} &
  \textbf{0.841 ± 0.176} &
  0.744 ± 0.182 &
  0.900 ± 0.197 &
  0.807 ± 0.165 &
  0.933 ± 0.130 &
  0.867 ± 0.116 \\ \cline{3-4} 
 &
  &
  \multicolumn{1}{l|}{} &
  LC-MIL-miNet &
  0.811 ± 0.181 &
  0.704 ± 0.195 &
  0.820 ± 0.198 &
  0.836 ± 0.202 &
  0.881 ± 0.120 &
  0.882 ± 0.124 \\ 
\multirow{-5}{*}{PAIP2020} &
  \multirow{-5}{*}{\shortstack{43.3 \% \\(± 13.0) \%}} &
  \multicolumn{1}{l|}{\multirow{-2}{*}{LC-MIL}} &
  LC-MIL-atten &
  0.823 ± 0.167 &
  0.718 ± 0.185 &
  0.833 ± 0.191 &
  0.848 ± 0.175 &
  0.886 ± 0.128 &
  0.882 ± 0.131 \\ \hline
\end{tabular}
\label{tab:overview-S-I-all}
\end{table*}

\begin{table*}[h]
\centering
\caption{Summary of refinement on S-II: Omit small lesions}
\begin{tabular}{|c|c|ll|c |l|c |c|c |l|}
\hline
Dataset &
  Tumor/tissue  &
  \multicolumn{2}{c|}{Method} &
  F1 &
  \multicolumn{1}{c|}{IoU} &
  PPV &
  TPR &
  TNR &
  \multicolumn{1}{c|}{NPV} \\ \hline
 &
  &
  \multicolumn{2}{l|}{Coarse annotations} &
  0.561 ± 0.136 &
  0.402 ± 0.129 &
  0.485 ± 0.147 &
  0.927 ± 0.177 &
  0.987 ± 0.012 &
  0.984 ± 0.020 \\ 
 &
  &
  \multicolumn{2}{l|}{DkNN} &
  0.597 ± 0.266 &
  0.471 ± 0.247 &
  0.774 ± 0.188 &
  0.861 ± 0.233 &
  0.998 ± 0.004 &
  0.981 ± 0.023 \\ 
 &
  &
  \multicolumn{2}{l|}{Rank Pruning} &
  0.758 ± 0.144 &
  0.630 ± 0.175 &
  0.844 ± 0.147 &
  0.869 ± 0.174 &
  0.998 ± 0.003 &
  0.986 ± 0.019 \\ \cline{3-4} 
 &
  &
  \multicolumn{1}{l|}{} &
  LC-MIL-miNet &
  0.840 ± 0.091 &
  0.735 ± 0.127 &
  0.773 ± 0.142 &
  0.922 ± 0.123 &
  0.996 ± 0.005, &
  0.993 ± 0.014 \\ 
\multirow{-5}{*}{\shortstack{CAME \\ LYON16  }} &
  \multirow{-5}{*}{\shortstack{29.7\% \\(± 17.0\%)}} &
  \multicolumn{1}{l|}{\multirow{-2}{*}{LC-MIL}} &
  LC-MIL-atten &
  \textbf{0.849 ± 0.084} &
  0.747 ± 0.117 &
  0.827 ± 0.137 &
  0.906 ± 0.119 &
  0.995 ± 0.008 &
  0.994 ± 0.012 \\ \hline
 &
  &
  \multicolumn{2}{l|}{Coarse annotations} &
  0.687 ± 0.168 &
  0.547 ± 0.186 &
  0.586 ± 0.179 &
  0.989 ± 0.029 &
  0.892 ± 0.076 &
  0.930 ± 0.093 \\ 
 &
  &
  \multicolumn{2}{l|}{DkNN} &
  0.722 ± 0.242 &
  0.612 ± 0.254 &
  0.713 ± 0.193 &
  0.971 ± 0.051 &
  0.947 ± 0.051 &
  0.929 ± 0.093 \\ 
 &
  &
  \multicolumn{2}{l|}{Rank Pruning} &
  0.768 ± 0.215 &
  0.666 ± 0.245 &
  0.766 ± 0.195 &
  0.955 ± 0.055 &
  0.962 ± 0.043 &
  0.931 ± 0.094 \\ \cline{3-4} 
 &
  &
  \multicolumn{1}{l|}{} &
  LC-MIL-miNet &
  \textbf{0.842 ± 0.119} &
  0.743 ± 0.165 &
  0.747 ± 0.208 &
  0.961 ± 0.053 &
  0.955 ± 0.040 &
  0.956 ± 0.069 \\ 
\multirow{-5}{*}{PAIP2019} &
  \multirow{-5}{*}{\shortstack{38.0\% \\(± 18.9\%)}} &
  \multicolumn{1}{l|}{\multirow{-2}{*}{LC-MIL}} &
  LC-MIL-atten &
  0.834 ± 0.128 &
  0.734 ± 0.173 &
  0.781 ± 0.219 &
  0.934 ± 0.077 &
  0.951 ± 0.049 &
  0.955 ± 0.067 \\ \hline
 &
  &
  \multicolumn{2}{l|}{Coarse annotations} &
  0.570 ± 0.076 &
  0.403 ± 0.070 &
  0.600 ± 0.130 &
  0.979 ± 0.045 &
  0.961 ± 0.030 &
  0.918 ± 0.058 \\ 
 &
  &
  \multicolumn{2}{l|}{DkNN} &
  0.548 ± 0.118 &
  0.386 ± 0.104 &
  0.834 ± 0.116 &
  0.968 ± 0.048 &
  0.991 ± 0.010 &
  0.909 ± 0.058 \\ 
 &
  &
  \multicolumn{2}{l|}{Rank Pruning} &
  0.607 ± 0.147 &
  0.452 ± 0.152 &
  0.867 ± 0.101 &
  0.901 ± 0.129 &
  0.993 ± 0.009 &
  0.916 ± 0.062 \\ \cline{3-4} 
 &
  &
  \multicolumn{1}{l|}{} &
  LC-MIL-miNet &
  \textbf{0.786 ± 0.106} &
  0.659 ± 0.138 &
  0.832 ± 0.115 &
  0.923 ± 0.057 &
  0.985 ± 0.014 &
  0.950 ± 0.055 \\ 
\multirow{-5}{*}{PAIP2020} &
  \multirow{-5}{*}{\shortstack{43.3\% \\(± 13.0\%)}} &
  \multicolumn{1}{l|}{\multirow{-2}{*}{LC-MIL}} &
  LC-MIL-atten &
  0.718 ± 0.138 &
  0.576 ± 0.157 &
  0.884 ± 0.100 &
  0.875 ± 0.075 &
  0.988 ± 0.014 &
  0.933 ± 0.064 \\ \hline
\end{tabular}
\label{tab:overview-S-II-all}
\end{table*}

\begin{table*}[h]
\centering
\caption{Summary of refinement on hand-drawn annotations}
\begin{tabular}{|c|c|ll|l|l|c|c|c|l|}
\hline
Dataset &
  Tumor/tissue &
  \multicolumn{2}{c|}{Method} &
  \multicolumn{1}{c|}{F1} &
  \multicolumn{1}{c|}{IoU} &
  PPV &
  TPR &
  TNR &
  \multicolumn{1}{c|}{NPV} \\ \hline
 &
  &
  \multicolumn{2}{l|}{Coarse annotations} &
  0.682 ±0.149 &
  0.537 ± 0.173 &
  0.656 ±0.219 &
  0.830 ±0.208 &
  0.980 ± 0.034 &
  0.992 ± 0.015 \\ 
 &
  &
  \multicolumn{2}{l|}{DkNN} &
  0.756 ±0.182 &
  0.639 ± 0.214 &
  0.832 ±0.167 &
  0.776 ±0.252 &
  0.992 ± 0.020 &
  0.990 ± 0.017 \\ 
 &
  &
  \multicolumn{2}{l|}{Rank Pruning} &
  0.830 ±0.121 &
  0.725 ± 0.154 &
  0.868 ±0.131 &
  0.836 ±0.168 &
  0.995 ± 0.009 &
  0.993 ± 0.013 \\ \cline{3-4} 
 &
  &
  \multicolumn{1}{l|}{} &
  LC-MIL-miNet &
  0.840 ±0.096 &
  0.736 ± 0.136 &
  0.819 ±0.141 &
  0.898 ±0.129 &
  0.994 ± 0.010 &
  0.994 ± 0.014 \\ 
\multirow{-5}{*}{\shortstack{CAME\\LYON16}} &
  \multirow{-5}{*}{\shortstack{29.7 \%\\ (± 17.0 \%)}} &
  \multicolumn{1}{l|}{\multirow{-2}{*}{LC-MIL}} &
  LC-MIL-atten &
  \textbf{0.848 ±0.096} &
  0.748 ± 0.133 &
  0.828 ±0.140 &
  0.901 ±0.114 &
  0.994 ± 0.012 &
  0.995 ± 0.012 \\ \hline
 &
  &
  \multicolumn{2}{l|}{Coarse annotations} &
  0.742 ±0.182 &
  0.618 ± 0.195 &
  0.647 ±0.202 &
  0.934 ±0.174 &
  0.848 ± 0.108 &
  0.982 ± 0.039 \\ 
 &
  &
  \multicolumn{2}{l|}{DkNN} &
  0.808 ±0.181 &
  0.708 ± 0.203 &
  0.753 ±0.202 &
  0.920 ±0.179 &
  0.912 ± 0.082 &
  0.979 ± 0.040 \\ 
 &
  &
  \multicolumn{2}{l|}{Rank Pruning} &
  0.830 ±0.168 &
  0.737 ± 0.197 &
  0.799 ±0.194 &
  0.913 ±0.155 &
  0.926 ± 0.086 &
  0.975 ± 0.044 \\ \cline{3-4} 
 &
  &
  \multicolumn{1}{l|}{} &
  LC-MIL-miNet &
  \textbf{0.833 ±0.167} &
  0.741 ± 0.197 &
  0.785 ±0.202 &
  0.932 ±0.131 &
  0.933 ± 0.070 &
  0.978 ± 0.030 \\ 
\multirow{-5}{*}{PAIP2019} &
  \multirow{-5}{*}{\shortstack{38.0 \% \\(± 18.9 \%)}} &
  \multicolumn{1}{l|}{\multirow{-2}{*}{LC-MIL}} &
  LC-MIL-atten &
  \textbf{0.833 ±0.163} &
  0.742 ± 0.203 &
  0.805 ±0.213 &
  0.919 ±0.114 &
  0.938 ± 0.075 &
  0.971 ± 0.034 \\ \hline
 &
  &
  \multicolumn{2}{l|}{Coarse annotations} &
  0.796 ± 0.114 &
  0.675 ± 0.147 &
  0.705 ± 0.152 &
  0.951 ± 0.092 &
  0.925 ± 0.075 &
  0.990 ± 0.022 \\ 
 &
  &
  \multicolumn{2}{l|}{DkNN} &
  \textbf{0.898 ± 0.095} &
  0.826 ± 0.126 &
  0.882 ± 0.105 &
  0.936 ± 0.112 &
  0.977 ± 0.034 &
  0.990 ± 0.022 \\ 
 &
  &
  \multicolumn{2}{l|}{Rank Pruning} &
  0.867 ± 0.110 &
  0.779 ± 0.141 &
  0.908 ± 0.091 &
  0.853 ± 0.152 &
  0.984 ± 0.027 &
  0.978 ± 0.028 \\ \cline{3-4} 
 &
  &
  \multicolumn{1}{l|}{} &
  LC-MIL-miNet &
  0.869 ± 0.081 &
  0.776 ± 0.115 &
  0.860 ± 0.112 &
  0.897 ± 0.103 &
  0.979 ± 0.020 &
  0.981 ± 0.027 \\ 
\multirow{-5}{*}{PAIP2020} &
  \multirow{-5}{*}{\shortstack{43.3 \% \\(± 13.0 \%)}} &
  \multicolumn{1}{l|}{\multirow{-2}{*}{LC-MIL}} &
  LC-MIL-atten &
  0.863 ± 0.077 &
  0.766 ± 0.111 &
  0.902 ± 0.097 &
  0.844 ± 0.114 &
  0.986 ± 0.019 &
  0.971 ± 0.035 \\ \hline
\end{tabular}
\label{tab:overview-real-world-all}
\end{table*}

\begin{figure*}[h]
\centerline{\includegraphics[width=\textwidth]{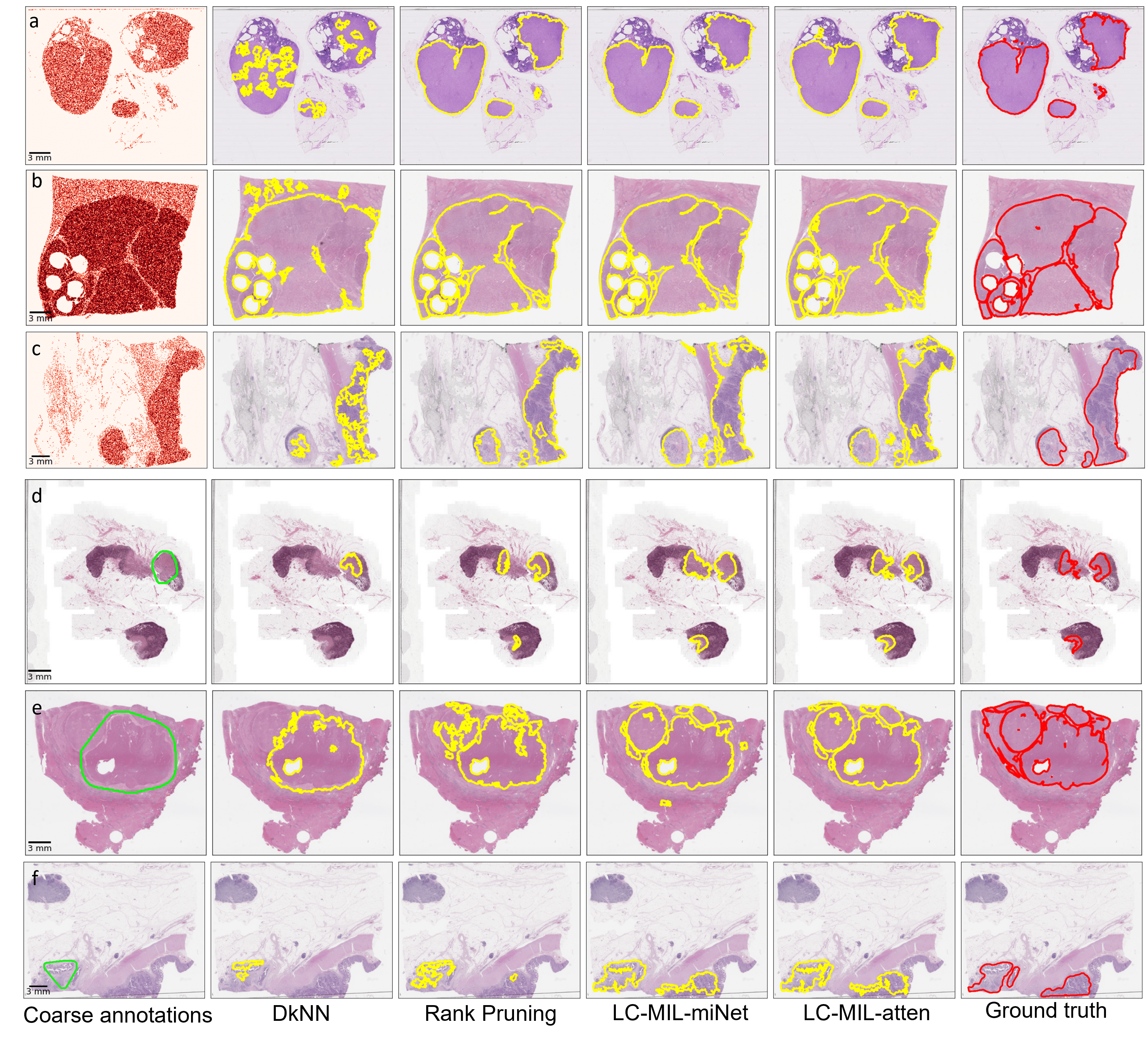}}
\caption{\textbf{Supplementary examples of simulated coarse annotations and refinement.}  (a-c) S-I: synthetic coarse annotations generated by uniformly flipping patches; (d-f): S-II: synthetic coarse annotations generated by omitting small lesions. Among those examples, (a) and (d) are from CAMELYON16, (b) and (e) are from PAIP2019, (c) and (f) are from PAIP2020. From left to right, six versions (coarse annotations (1st column, lime lines); predicted contours using DkNN (2nd column, yellow lines); predicted contours using Rank Pruning (3rd column, yellow lines); predicted contours using LC-MIL-miNet (4th column, yellow lines)); predicted contours using LC-MIL-atten (5th column, yellow lines)); ground truth (6th column, red lines)) of cancerous regions contours are shown.}
\label{fig:examples-simulated-supp}
\end{figure*}

\begin{figure*}[h]
\centerline{\includegraphics[width=\textwidth]{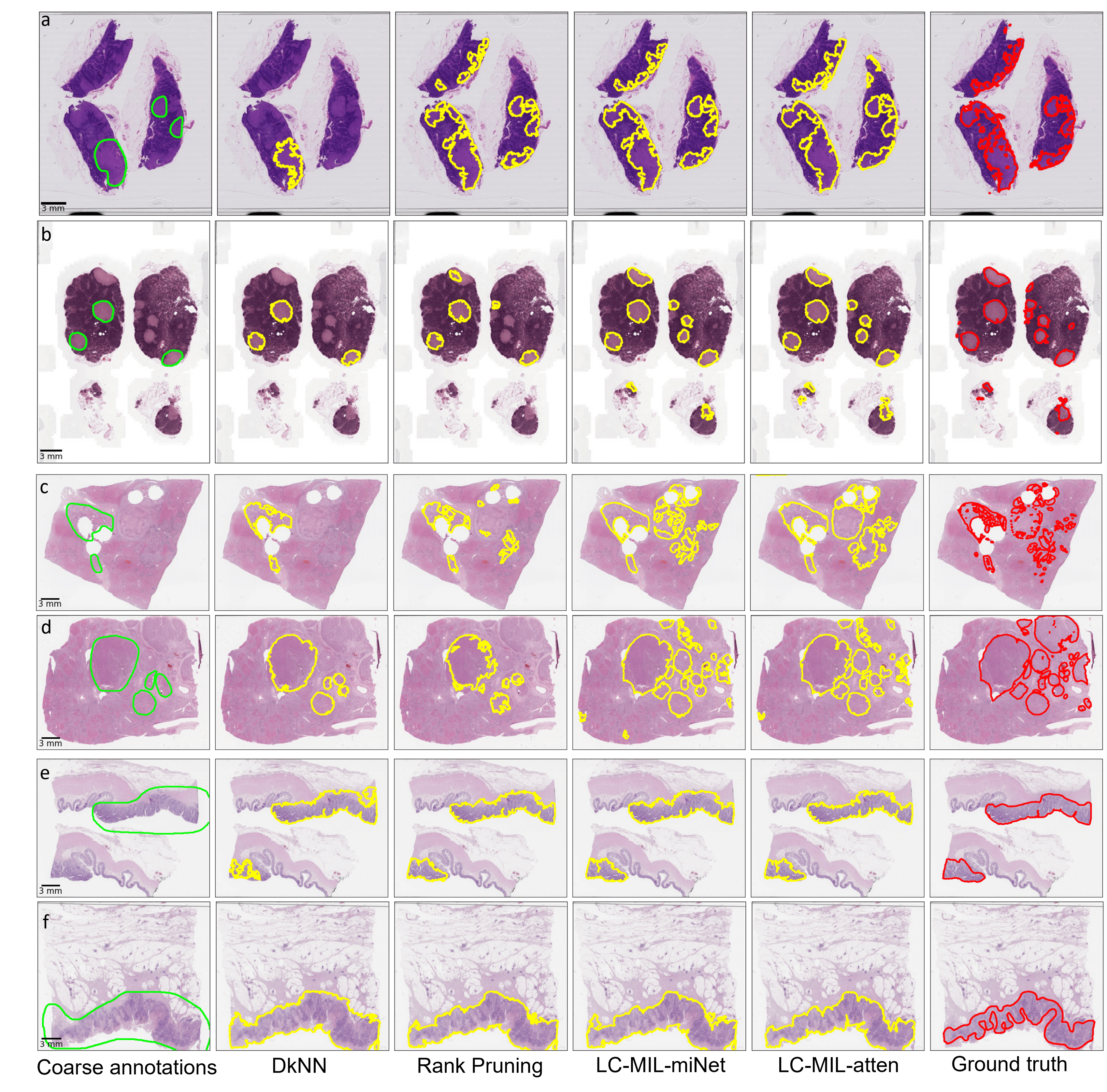}}
\caption{\textbf{Supplementary examples of hand-drawn coarse annotations and refinement.} Among those examples, (a) and (b) are from CAMELYON16, (c) and (d) are from PAIP2019, (e) and (f) are from PAIP2020. From left to right, six versions (coarse annotations (1st column, lime lines); predicted contours using DkNN (2nd column, yellow lines); predicted contours using Rank Pruning (3rd column, yellow lines); predicted contours using LC-MIL-miNet (4th column, yellow lines)); predicted contours using LC-MIL-atten (5th column, yellow lines)); ground truth (6th column, red lines)) of cancerous regions contours are shown.} 
\label{fig:examples-real-world-supp}
\end{figure*}

\begin{figure*}[h]
\centering
\centerline{\includegraphics[width=\textwidth]{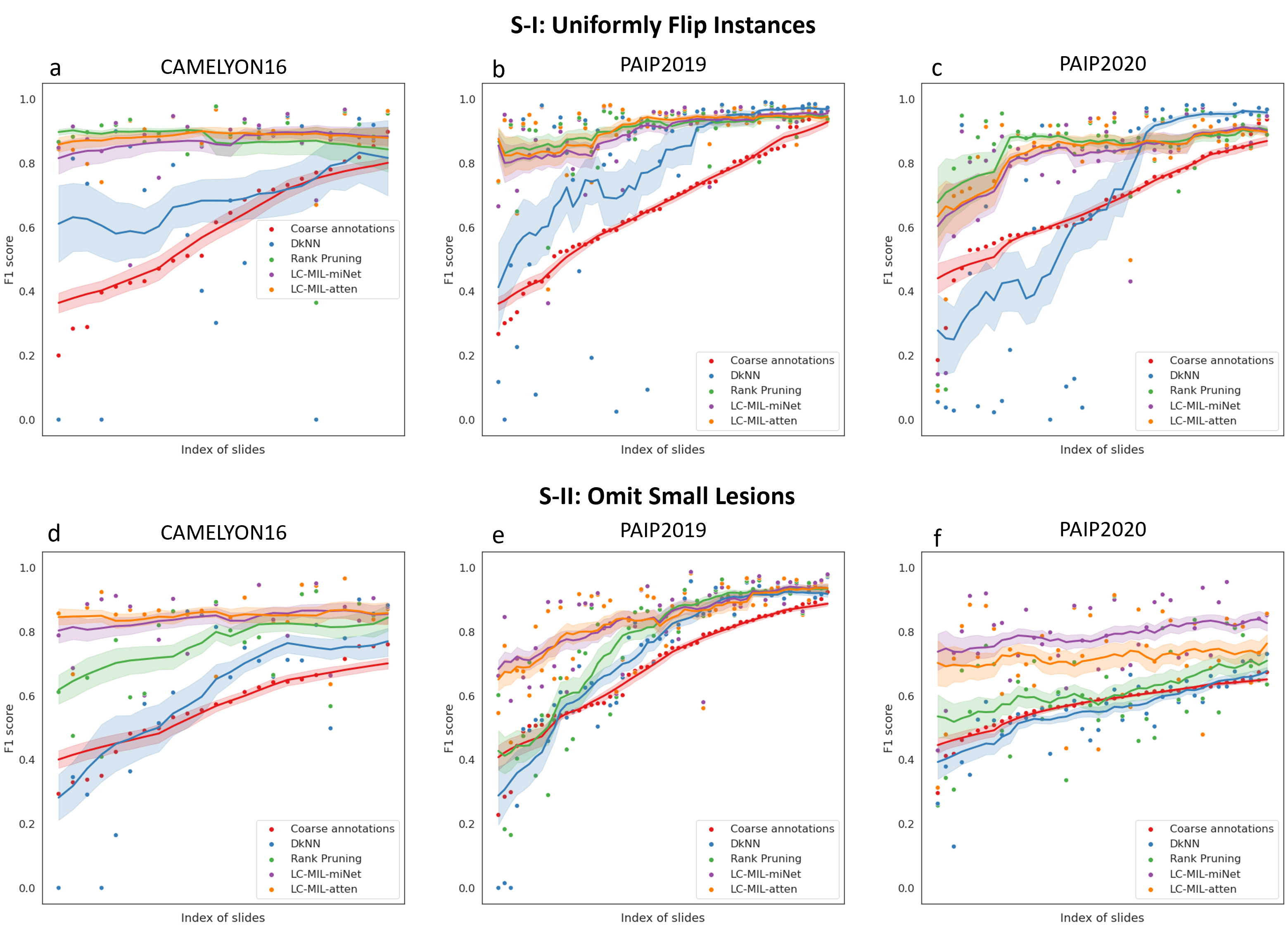}}
\caption{\textbf{Refinement performance under different noise conditions (synthetic annotations).} F1 scores of coarse annotations (red), refined annotations by DkNN (blue), Rank Pruning (green), LC-MIL-miNet (purple), and LC-MIL-atten (organge), with slides sorted by F1 scores of their coarse annotations . The moving average (solid lines), where window size $k=15$, and the 68\% confidence interval for $k$ observations (shaded areas) are also shown.}
\label{fig:scatterplot-simulated}
\end{figure*}

\begin{figure*}[h]
\centering
\centerline{\includegraphics[width=\textwidth]{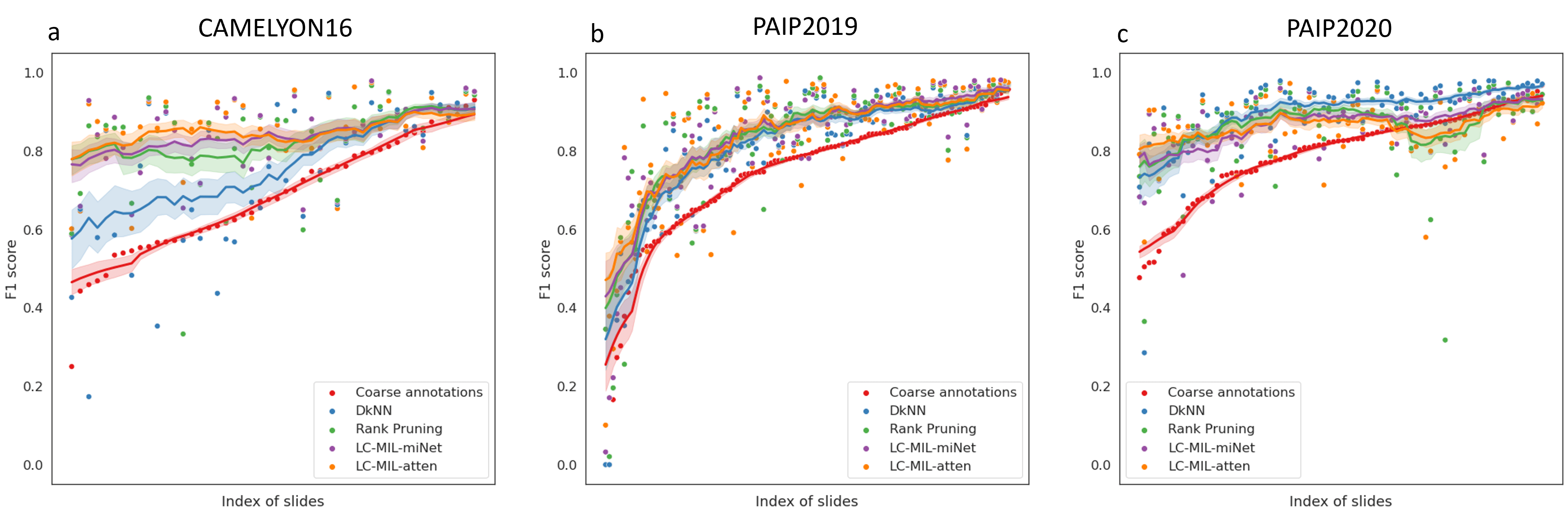}}
\caption{\textbf{Refinement performance under different noise conditions (hand-drawn annotations).} F1 scores of coarse annotations and refined annotations (using different methods), with slides are sorted by F1 scores of their coarse annotations. The moving average (solid lines), where window size $k=15$, and the 68\% confidence interval for $k$ observations (shaded areas) are also shown.}
\label{fig:scatterplot-real-world}
\end{figure*}

\end{document}